%% file: main.tex
\documentclass[10pt,twocolumn,letterpaper]{article}

\usepackage{wacv}              % To produce the CAMERA-READY version
\usepackage{algorithm}
\usepackage{algpseudocode}
\usepackage[accsupp]{axessibility}  % Improves PDF readability for those with disabilities.


\usepackage{multirow}
\usepackage{array,tabularx,booktabs}
\usepackage{graphicx}
\usepackage{array}
\usepackage{booktabs}
\usepackage{makecell}

\newcolumntype{L}{>{\centering\arraybackslash}p{\dimexpr0.15\textwidth-2\tabcolsep}}
\newcolumntype{C}{>{\centering\arraybackslash}p{\dimexpr0.283\textwidth-2\tabcolsep}} % ~3xC fits

\newlength{\RefH}
\newlength{\TopH}
\newlength{\BotH}
\newlength{\LabH}

\newcommand{\LabelBox}[1]{\vbox to \LabH{\vfil{\scriptsize #1}\vfil}}

\newcommand{\RefCell}[1]{%
  \parbox[t]{\linewidth}{\centering
    \includegraphics[width=\linewidth,height=\RefH,keepaspectratio]{#1}%
  }%
}
\newcommand{\StackTwo}[2]{%
  \parbox[t]{\linewidth}{\centering
    \includegraphics[width=\linewidth,height=\TopH,keepaspectratio]{#1}\par
    \LabelBox{Baseline}\vspace{2pt}
    \includegraphics[width=\linewidth,height=\BotH,keepaspectratio]{#2}\par
    \LabelBox{Pipeline}%
  }%
}

\newcommand{\PromptRowFour}[1]{%
  \multicolumn{4}{p{\dimexpr\textwidth-2\tabcolsep}}{\footnotesize\textbf{Prompt:}~#1} \\
}

\newcolumntype{J}{>{\centering\arraybackslash}p{0.14\textwidth}}
\newcolumntype{K}{>{\centering\arraybackslash}p{0.35\textwidth}}

\newlength{\RefHthree}
\newlength{\CellHthree}

\newcommand{\PromptRow}[1]{%
  \PromptRowFour{#1}% Use 4-column version by default
}

\definecolor{wacvblue}{rgb}{0.21,0.49,0.74}
\usepackage[pagebackref,breaklinks,colorlinks,allcolors=wacvblue]{hyperref}

\def\wacvPaperID{3161} % *** Enter the WACV Paper ID here
\def\confName{WACV}
\def\confYear{2026}

\title{From Prompt to Production: Automating Brand-Safe Marketing Imagery with Text-to-Image Models}

\author{
Parmida Atighehchian \and
Henry Wang \and
Andrei Kapustin \and
Boris Lerner \and
Tiancheng Jiang \and
Taylor Jensen \quad Negin Sokhandan\\
Amazon Web Services \\
{\tt\small \{patigheh, yuanhenw, kapustan, blerner, tjiangg, tdjensen, ngnsl\}@amazon.com}
}

\begin{document}
\maketitle
\input{sec/0_abstract}    
\input{sec/1_intro}

\input{sec/2_relatedwork}
\input{sec/3_method}
\input{sec/4_experiments}
\input{sec/5_conclusion}
{
    \small
    \bibliographystyle{ieeenat_fullname}
    \bibliography{main}
}
\end{document}

\input{sec/6_supplementary}

%% file: sec/0_abstract.tex
\begin{abstract}
Text-to-image models have made significant strides, producing impressive results in generating images from textual descriptions. However, creating a scalable pipeline for deploying these models in production remains a challenge. Achieving the right balance between automation and human feedback is critical to maintain both scale and quality. While automation can handle large volumes, human oversight is still an essential component to ensure that the generated images meet the desired standards and are aligned with the creative vision. This paper presents a new pipeline that offers a fully automated, scalable solution for generating marketing images of commercial products using text-to-image models. The proposed system maintains the quality and fidelity of images, while also introducing sufficient creative variation to adhere to marketing guidelines. By streamlining this process, we ensure a seamless blend of efficiency and human oversight, achieving a $30.77\%$ increase in marketing object fidelity using DINOV2 and a $52.00\%$ increase in human preference over the generated outcome.
\end{abstract}

%% file: sec/1_intro.tex
\section{Introduction}
\label{sec:introduction}

The rapid advancement of generative artificial intelligence has transformed content creation workflows across industries, with marketing and advertising emerging as particularly active domains~\cite{dalle2022, stablediff2022}. Recent breakthroughs in text-to-image generation models have enabled high-quality visual creation from natural language descriptions, leading to widespread experimentation in commercial applications~\cite{stablediff2022,dalle2022,imagen2022,podell2023sdxl,flux2024}. Despite these advances, integrating such models into production marketing workflows remains challenging, limiting their practical deployment at scale.

The transition from experimental AI-generated content to production-ready marketing materials reveals critical gaps in current approaches. While text-to-image models excel at producing visually appealing content, they struggle to consistently meet the stringent requirements of professional marketing applications of a commercial product (like a shoe), where adherence to brand guidelines, product fidelity, and scalable workflows is paramount. Here, brand guidelines mean accurate representation of product-specific features (e.g., colorways, logos, stitching). Efforts to address these limitations through architectural innovations such as ControlNet~\cite{controlnet2023}, adapters~\cite{t2iAdapter}, and IP-Adapters~\cite{ipAdapter} provide improved control but fall short of delivering the comprehensive solution required for enterprise use.

We identify three fundamental challenges preventing current generative AI systems from achieving production-grade marketing content generation:

\textbf{Alignment with brand guidelines}: Marketing content must adhere to brand specifications (e.g., colors, background features) while preserving product characteristics without alteration. The inherent variability of generative models can compromise brand consistency and product accuracy. Control mechanisms such as ControlNet~\cite{controlnet2023} and fine-tuning methods like AnyDoor~\cite{anydoor} offer structural guidance. However, they cannot guarantee the pixel-level fidelity required for marketing campaigns, where even small deviations can compromise brand integrity. Outpainting models aim to maintain reference object fidelity, but common failure cases include duplicating objects, hallucinations (i.e. generating spurious elements), or ignoring the object entirely. 

\textbf{Predictable outcome and quality consistency}: A common solution for campaign creation is to incorporate outpainting. However, seamlessly integrating products into scenes remains challenging and varies significantly based on product and scene characteristics. Skilled creators can iteratively refine generations using visual heuristics, prompt tuning, and trial and error, but creators rarely achieve quality outcomes on the first attempt.

\textbf{Scalability and workflow integration}: Production workflows demand consistent, high-throughput content generation with minimal human oversight. Current approaches require repeated manual iteration, prompt engineering, and quality checks, creating bottlenecks that hinder scalability. Reliance on constant human input makes these systems impractical for enterprise campaigns that require hundreds of variations.

To address these challenges, we present a novel end-to-end automated system for generating composite marketing images that respect brand guidelines while enabling scalable and robust workflows. Our approach creates a production-ready workflow by deferring human validation to the final stage and automating all intermediate generations and quality assessments through an agentic pipeline.

Our system introduces several key innovations: (1) a structured prompt decomposition mechanism that converts complex marketing requirements into machine-readable specifications, (2) an intelligent asset retrieval system with specialized guardrails for content types, (3) a rich caption generator and a multi-modal composition planer that determines optimal product placement and scaling, (4) a grid search approach for marketing image variants, and (5) an intelligent quality control pipeline that evaluates content across multiple dimensions without human intervention. By treating the core generation module as plug-and-play, the system adapts to diverse industry needs at scale.

Section~\ref{sec:method} details each pipeline module. Our experiments show that the proposed approach achieves significant gains in generation quality compared to existing methods, while maintaining strict adherence to brand guidelines and composition standards. The system handles diverse marketing scenarios and provides the scalability required for enterprise deployment.

%%%% This is a bit of a repeat and also too much emphasis on prompt decomposition where we don't even test in the paper.%%%%%
% The primary contributions of this work are:

% \begin{itemize}
%     \item \textbf{End-to-End Automated Pipeline with Multi-Stage Quality Control}: We develop a comprehensive automation framework that eliminates intermediate human validation steps while maintaining production-grade output quality through systematic multi-modal assessment. Our pipeline processes complex marketing prompts and delivers filtered, ranked results without manual intervention.
    
%     \item \textbf{Adaptive Quality Guardrails for Brand Consistency}: We introduce a novel agentic multimodal guardrail system that evaluates generated content across four critical dimensions: caption alignment, product uniqueness, physical realism, and lighting consistency. This approach enables brands to maintain consistency through configurable text-based criteria while preserving product integrity throughout the generation process.
    
%     \item \textbf{Structured Prompt Decomposition for Complex Marketing Requirements}: We propose a systematic approach to parsing complex marketing prompts into structured components (primary products, accessories, backgrounds, themes, and tactics), enabling precise asset retrieval and composition planning. This decomposition bridges the gap between natural language marketing requirements and machine-executable generation parameters.
% \end{itemize}

%% file: sec/2_relatedwork.tex
\section{Related Work}
\label{sec:relatedwork}
While various research efforts have explored improving control over text-to-image generation, to our knowledge, no prior work has proposed a scalable workflow with comprehensive testing for production-ready deployment. We therefore review the most relevant approaches to object compositing (the core component of our system), particularly in the context of marketing applications.  

\subsection{Fine-tuned Object Insertion and Harmonization}
ICLight~\cite{iclight} is a diffusion-based framework for illumination harmonization and editing that operates on uncurated, in-the-wild data. ICLight enforces light transport consistency, enabling physically plausible relighting and compositional edits. However, ICLight can alter product textures and colors, creating a challenge for brand-sensitive marketing content.  
AnyDoor~\cite{anydoor} enables zero-shot object-level customization by teleporting a target object into a new scene at a user-specified location without per-object fine-tuning. It balances identity preservation with local variations in lighting and pose.
ObjectStitch~\cite{objectStich} similarly performs object insertion using conditional diffusion, with adjustments such as geometry correction, color harmonization, and shadow synthesis. Despite these capabilities, AnyDoor and ObjectStitch struggle to maintain strict object fidelity, limiting their reliability in marketing scenarios.  

\subsection{Training-free Insertion}
Insert Diffusion~\cite{insertDiff} proposes a training-free approach for inserting objects into scenes. However, the method requires manual positioning and scaling, and a second diffusion pass, which can cause unwanted pixel-level manipulation.

\subsection{Outpainting}
Inpaint Anything~\cite{in-paint-anything} introduces a modular inpainting framework built around a “clicking and filling” paradigm: objects can be removed, replaced, or preserved while modifying the background based on user prompts. While flexible, the method relies on manual object selection, preventing full automation. More broadly, outpainting models often hallucinate artifacts near objects or replicate them unnaturally within the scene, further complicating their use in marketing image generation.  

Overall, while prior methods advance object compositing and harmonization, they fall short of generating brand-consistent, production-ready marketing content without extensive manual refinement. Section~\ref{sec:method} details how our pipeline overcomes these limitations through systematic generation and artifact filtering.  

% Magic Insert \cite{ruiz2024magic} proposes a style‐aware drag-and-drop method for inserting subjects from one image into a target image with a different style, while preserving physical plausibility and stylistic consistency. This is acheived by fine-tuning a pretrained text-to-image diffusion model using LoRA and subject-specific text tokens, combined with a CLIP-based encoding of the target style. For object insertion, the method employs Bootstrapped Domain Adaptation to adapt a photorealistic insertion model to diverse artistic styles.

%% file: sec/3_method.tex
\section{Method}
\label{sec:method}
We present an automated system for generating high-quality composite marketing images from natural language prompts. Our approach translates marketing requirements into visually compelling product compositions via a four-stage pipeline: structured prompt analysis, intelligent asset retrieval, composition planning and variation generation, and multi-modal quality assessment.

\subsection{System Overview}
% \afterpage{
\begin{figure*}[t]  % note the *
    \centering
    \includegraphics[width=0.9\textwidth]{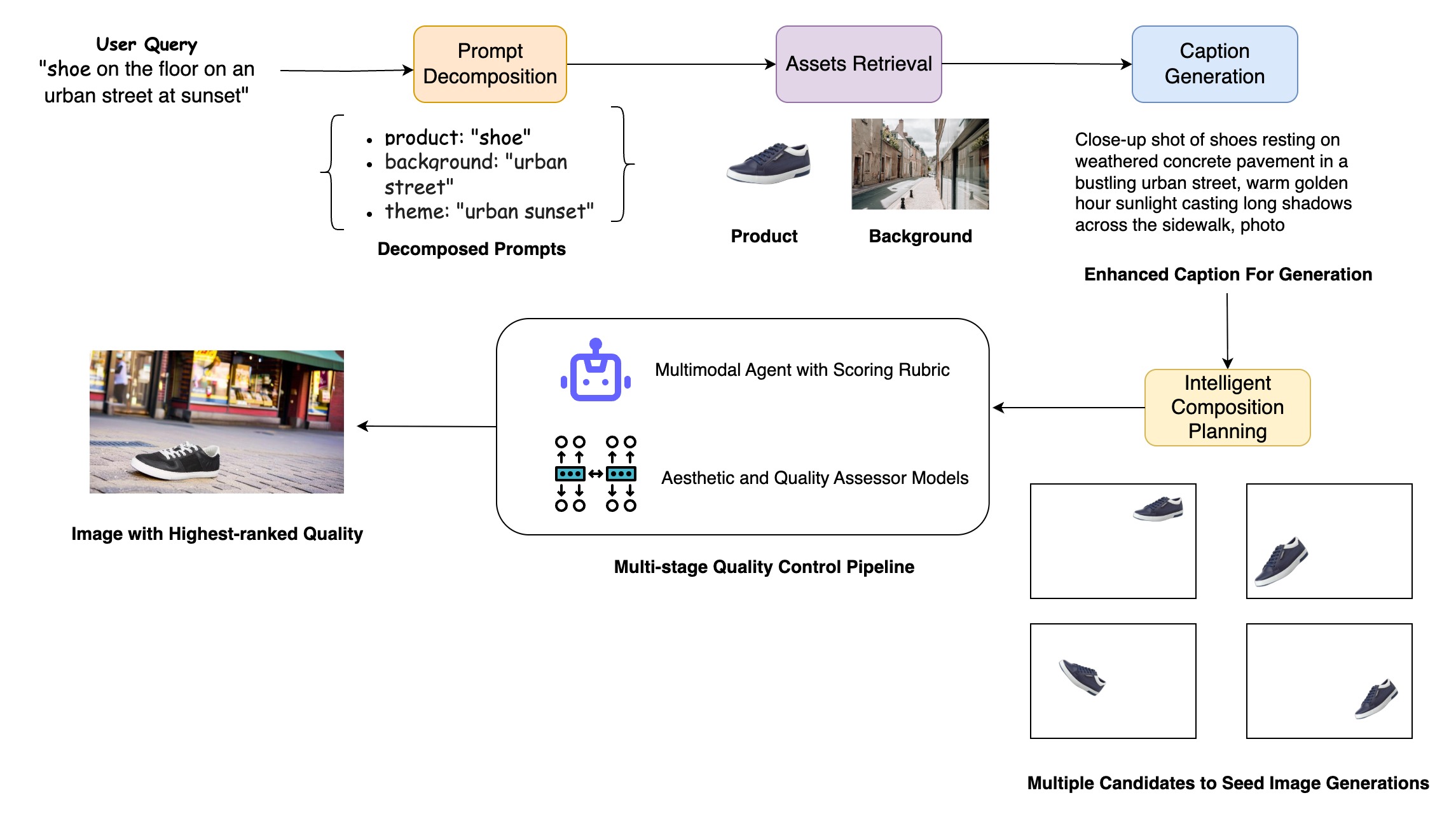}
    \caption{Overview of our pipeline. The pipeline starts with decomposing the user query. It then retrieves relevant visual assets, and rewrites the initial prompts for optimized image generation quality. The visual assets are scaled and positioned at various positions of the canvas to maximize generation diversity. After generation, candidate images are filtered and ranked based on rubric scores and aesthetic scores.}
    \label{fig:system_overview}
\end{figure*}
% }

Given a marketing prompt such as \textit{"Shoe on the floor on an urban street at sunset"}, the system automatically: (1) decomposes the prompt into structured components, (2) retrieves relevant visual assets, (3) determines optimal composition parameters using a caption generator and a multi-modal advisor, generates multiple image variations, and (4) performs comprehensive quality filtering to select the top-$k$ final images. This automation minimizes manual intervention while maintaining professional-grade quality (Figure \ref{fig:system_overview}.)

\subsection{Structured Prompt Decomposition}
Complex marketing prompts often contain multiple semantic elements and implicit requirements. We address this by converting natural language prompts into machine-readable specifications using a large language model.
Our decomposition module extracts three key components: (1) \textbf{primary product} - the main item being promoted, (2) \textbf{background elements} - environmental context and scene descriptors, (3) \textbf{theme} - the promotional context or occasion. This structured representation ensures all specified elements are considered during asset retrieval and composition planning. The decomposition process uses few-shot prompting with predefined output schemas to maintain consistency across diverse input formats.
\subsection{Multi-modal Asset Retrieval with Guardrails}
Precise asset selection is critical for composite image generation. We employ a two-tier retrieval system with specialized guardrails for different asset types.
\subsubsection{Product Asset Retrieval}
Product assets are retrieved using similarity-based dense vector representations. Candidates are filtered by a similarity threshold $\tau_p = 0.39$, empirically optimized to balance precision and recall, ensuring visually and semantically appropriate products.
\subsubsection{Background Asset Retrieval with LLM Validation}
Background selection is more complex due to subjective scene appropriateness. While similarity scores provide an initial ranking, they often miss nuanced contextual requirements. After retrieving the top-$k$ most relevant matches, we employ an LLM-based validation step that evaluates semantic alignment between each candidate background $B_i$ and prompt $P$:

\begin{equation}
R(B_i, P) = \text{LLM}_{\text{validator}}(B_i, P) \in {0, 1}
\end{equation}

This step ensures robust selection by considering context, theme, and visual compatibility.

\subsection{Caption Generator}
The caption generator is an LLM agent prompted to produce descriptive captions that guide image generation and composition planning. It considers the background, primary product, and theme to emphasize the product while positioning other elements appropriately. If a relevant background is found, the caption generator uses the image to enrich the text with additional visual features. Otherwise, it relies on the user query. This caption not only is used for image generation but also plays an important role in the composition planning stage.

\subsection{Intelligent Composition Planning}
\label{sec:composit-planning}
Our composition module determines optimal object placement, scaling, and spatial relationships using multi-modal analysis and grid search.
\subsubsection{Multi-modal Composition Analysis}
A critical design decision is the ability to continue with the image generation even if a background match is not found.
Using either an empty canvas or a resized background, the multi-modal advisor leverages the generated caption to recommend scaling factors for the product in the scene:

Given a background-filled or an empty canvas $C$, the product object $O$, and the generated caption $Ca$ our composition advisor analyzes both visual and textual elements to determine optimal scaling factors:

\begin{equation}
S = \{s_w, s_h\} = f_{\text{advisor}}(C, O, Ca)
\end{equation}

where $s_w$ and $s_h$ represent the recommended width and height scaling factors relative to the canvas dimensions.

\subsubsection{Variation Generation Strategy}
To maximize creative diversity, products are systematically placed across canvas regions (left, center, right thirds) with rotation angles $\theta \in \{0^\circ, 15^\circ, 345^\circ\}$. This ensures broad exploration of the composition space while maintaining visual coherence.

\subsubsection{Scene Generation}
For marketing campaign generation, maintaining product fidelity is crucial. Our scene generation module is designed as a plug-and-play component that works with any image generation model accepting reference images and captions. For marketing applications, the generative model must preserve product fidelity while properly integrating objects into scenes. Given these base performance requirements, any outpainting, insertion and harmonization, or object composition model is suitable.

This module takes a product-on-canvas composition and extend the background while preserving visual integrity. Although the original background cannot be maintained, our feature-rich caption generation injects key background characteristics back into the scene, enabling better product composition. This approach ensures seamless product integration with realistic lighting and perspective relationships.

The image variants encourage the model to sample diversely from its learned latent space. This systematic exploration of the latent space increases the probability of generating high-quality marketing images by leveraging the model's full generative capacity.

% It is noteworthy that while our pipeline is designed to make the existing generative models more robust and stable to be able to plug in a production setup, we are not improving the performance of the core image generative model. Instead, our pipeline pushes the model to sample from a larger part of the embedding space using our grid search approach and by setting the product to the right scale, we give model a better guidance. Together these component, combined with the enriched caption sets the model up for a better high performance hit rate. 

\subsection{Multi-Stage Quality Control Pipeline}
\label{sec:qualitycontrol}
Quality assessment evaluates semantic accuracy, visual realism, and aesthetic appeal. Our pipeline combines rule-based validation with learned metrics and is fully customizable for different use cases.
\subsubsection{Multi-modal Quality Controller}
Our primary quality filter employs a multi-modal Large Language Model that evaluates generated images across four primary critical dimensions:

\begin{enumerate}
\item \textbf{Caption Alignment}: Verification that all elements specified in the prompt are accurately depicted in the generated image.
\item \textbf{Product Uniqueness}: Confirmation that the scene contains exactly one instance of the product—a safeguard against common model hallucinations (e.g., unintended duplicates)
\item \textbf{Physical Realism}: Assessment of object placement consistency with real-world physics and spatial relationships.
\item \textbf{Lighting Consistency}: Evaluation of shadow and reflection accuracy relative to the scene's lighting conditions.
\end{enumerate}

Each criterion receives a binary score $c_i \in \{0, 1\}$. In most production setups, the overall score is computed as below:

\begin{equation}
G = \prod_{i=1}^{4} c_i
\end{equation}
This multiplicative formulation ensures that images failing any critical criterion are filtered out, maintaining high quality standards.
Where data is sufficient, we recommend using a weighted average, where the weights are optimized to the use case in hand. In Section \ref{sec:exp}, we use a hierarchical approach to relax the less critical guardrails for our setup. 
Since the generative model is probabilistic, this component is a way to trigger a retry if necessary in a production setup and have a safe fall back.

\subsubsection{Aesthetic and Semantic Quality Assessment}
For critical applications such as marketing image creation, we recommend maintaining human oversight while minimizing intervention to enable scalable deployment. To facilitate scaling, the number of final images presented for human
review is configurable through a hyperparameter $k$. When more than $k$ candidates pass the guardrail evaluation, they undergo additional quality assessment using complementary metrics:

 \textbf{Aesthetic Scoring}: We employ a pre-trained aesthetic quality model \cite{aesthetic_pred} that processes CLIP \cite{clip} visual features to predict aesthetic scores. Images below the set aesthetic threshold are filtered out.

\textbf{CLIP-based Semantic Alignment}: We optionally measure text-image alignment using CLIP similarity between generated images and prompt descriptions as an additional filter. The CLIP \cite{clip} score is computed as:

\begin{equation}
\text{CLIP}(I, T) = w \cdot \max(0, \cos(\phi(I), \psi(T)))
\end{equation}

where $\phi(I)$ and $\psi(T)$ are the CLIP encodings of image $I$ and text $T$, respectively, and $w = 2.5$ is a scaling factor.

To reduce the cost overhead and latency, we chose to incorporate metrics that need minimal computation power. However, each of the modules are fully customizable and can be adapted to different sets of requirements.
\subsubsection{Combined Quality Ranking}
Final image ranking combines aesthetic and semantic quality scores:

\begin{equation}
Q_{\text{final}} = \alpha \cdot Q_{\text{aesthetic}} + \beta \cdot Q_{\text{CLIP}}
\end{equation}

where $\alpha$ and $\beta$ are weighting parameters that can be adjusted based on application requirements. This multi-faceted approach ensures that selected images excel across both perceptual quality and semantic accuracy dimensions.

The complete pipeline processes multiple image variations and returns only those meeting all quality criteria, ranked by their combined quality scores.

%% file: sec/4_experiments.tex
\section{Experiments}
\label{sec:exp}
\subsection{Baseline}
\label{sec:baseline}

We select baseline candidates based on their ability to perform outpainting and harmonization using reference images and prompts. Below is the list of our selected candidates:
\begin{itemize}
    \item \textbf{ICLight \cite{iclight}}: ICLight represents state-of-the-art object-scene compositing with specialized fine-tuning for lighting and shadow adjustment. Unlike standard outpainting models that focus solely on semantic alignment, ICLight's illumination capabilities enable more natural product integration.

    \item \textbf{Inpaint Anything ~\cite{in-paint-anything}}: Replace Anything module from the Inpaint Anything framework segments the object of interest from the background. It then replaces the background with a text-guided inpainting model. This model requires an input to select the object for segmentation (a “clicking and filling” paradigm). In our image variants, the object location is determined. We use this prior placement knowledge to automate the object selection.

    \item \textbf{Amazon Nova Canvas ~\cite{langford2025amazon}}: Nova Canvas receives the product on an empty canvas. Additionally, it has capabilities to accept a mask image or a mask prompt to mask out the product and perform text guided outpainting on the scene. For simplicity, we use the mask-prompt in our experiments.
\end{itemize}

We compare our approach to these baselines along two verticals critical to our application domain: (1) Product Fidelity and (2) Scene Integration.

\subsection{Data}
We use the Amazon-Berkeley Objects (ABO) dataset \cite{collins2022abo} as the primary source for product images. The ABO dataset contains high-quality, multi-view images of real-world products across diverse categories with annotations. These product images are clean and have a white background which eliminates the need for any preprocessing. We select 21 distinct categories, ranging from clothing and accessories to household items to cover a diverse range of items. We sample a minimum of 10 images per category if there are a sufficient number of images to sample from. If the number of images in a category is less than 10, we include all of them in our test dataset. Our final sampled dataset consists of 206 images representing 21 different categories. 

To further increase our test sample size, we send each product type to a Visual Language Model (VLM) and ask the model to generate 5 rich, diverse and creative captions for scenes that the product can fit in. With 5 captions per product, we have a total of 1030 product-caption pairs for our experiments.  

\subsection{Experiment setup}
We evaluate all three baselines on the same set of 1,030 product–caption pairs. Each method requires a mask image specifying product placement. In real-world workflows, designers typically adjust product position and rotation through repeated trials until a visually coherent result is achieved. Our pipeline automates this process by placing the product at multiple positions and rotations on an empty canvas, enabling the model to generate diverse variations in a single batch. This increases the likelihood of obtaining a high-quality composition without manual intervention.

To ensure fair comparison, we restrict each baseline to a single fixed placement–rotation configuration, which represents a non-automated workflow where no iterative adjustment is performed. By contrasting one-shot baseline outputs with our multi-variation pipeline, we directly test the contribution of automated variation generation.

For pipeline evaluation, we initialize from ground-truth product–caption pairs, bypassing retrieval and captioning, and begin at the composition stage. The setup follows Section~\ref{sec:composit-planning}, using an empty canvas. The final pipeline outputs are selected through our quality assessment module. In this experiment, we follow algorithm~\ref{alg:find-best} to progressively relax the lower-priority rules if needed, while keeping rules 2 and 3 strict (see Section~\ref{sec:qualitycontrol} for the list of rules).

\begin{algorithm}[t]
\caption{Find images aligned with guidelines.}
\label{alg:find-best}
\begin{algorithmic}[1]
\State \textbf{Input:} A list of lists containing 4 binary $scores$
\State \textbf{Output:} A list of image indices matching the one of the best pattern, or $\emptyset$
\State $patterns \gets [[1,1,1,1], [0,1,1,1], [0,1,1,0]]$
\For{each $pattern$ in $patterns$}
    \State $indices \gets \{ i \mid scores[i] = pattern \}$
    \If{$indices \neq \emptyset$}
        \State \Return $indices$
    \EndIf
\EndFor
\State \Return $\emptyset$
\end{algorithmic}
\end{algorithm}

Figure~\ref{fig:baseline-vs-workflow-grid} shows a comparison of our pipeline generated images against the baseline models' generations (additional results can be found in Supplementary Material section 1.1, Figures 1, 2, and 3).

\subsection{Quantitative Results} 
We use the following metrics to evaluate performance across the three key axes:

\begin{itemize}
    \item \textbf{Product Fidelity}: We use DINOv2 ~\cite{oquab2023dinov2} features and MS-SSIM ~\cite{ms-ssim} to evaluate how well the original product is preserved in the generated image. To do so, one needs to isolate the generated product from the scene. We accomplish this by using Amazon Nova Pro~\cite{langford2025amazon} as an image analyzer agent to predict bounding box coordinates around the product, which are then passed with the image to SAM2~\cite{ravi2024sam} for segmentation. We then compare the extracted object with the ground-truth product using the cosine similarity of DINOv2 features and calculated MS-SSIM values. DINOv2 similarity score captures both semantic and fine-grained visual similarities, while MS-SSIM provides a strong signal of structural fidelity, detecting distortions or visual degradation. Combined, these two metrics offer a robust measure of product fidelity in the generated scenes.
    \item \textbf{Overall text to image alignment}: We compute the HPSv2 \cite{wu2023human} to measure the overall quality of the image and alignment with text. While not a direct measure of quality for images, HPSv2 correlates strongly with human visual preference.
\end{itemize}

Tables~\ref{tab:merged} report results for each baseline and their integration into our pipeline. Across all models, our pipeline consistently improves both structural similarity and DINOv2 scores, supporting our claim that object size and placement play a critical role in outpainting quality. Although we observe a slight drop in HPSv2 for ICLight, our human evaluation indicates a strong preference for pipeline-generated images. This drop is partly explained by our pipeline’s design: when no images pass the quality check, the output defaults to a score of $0$. This effect is more pronounced in models like ICLight, where the baseline often alters an object’s texture or color. In such cases, the baseline still receives a score for its output, even if it deviates from fidelity requirements, while the pipeline registers no result (see Limitations ~\ref{sec:limit}). Consequently, more test cases yield zero outputs on the pipeline side, lowering the average HPSv2. In contrast, we observe a slight improvement in HPSv2 for Replace Anything. This improvement stems from selecting more effective product placement and rotation, which enables better integration. The baseline Replace Anything frequently hallucinates around products, whereas our pipeline mitigates this by providing stronger initialization. 

\begin{table*}[t]
  \captionsetup{justification=raggedright, singlelinecheck=false, font=small} % optional
  \centering
  \setlength{\tabcolsep}{5pt}             % column padding
  \renewcommand{\arraystretch}{1.12}      % row height
  \small                                  % readable size (not tiny)

  \caption{Comparison of metrics across three models with and without the proposed pipeline. Values are mean $\pm$ std.}
  \label{tab:merged}

  \begin{tabularx}{\textwidth}{l *{6}{>{\centering\arraybackslash}X}}
    \toprule
    & \multicolumn{2}{c}{\textbf{IClight}} 
    & \multicolumn{2}{c}{\textbf{Nova Canvas}} 
    & \multicolumn{2}{c}{\textbf{Replace Anything}} \\
    \cmidrule(lr){2-3}\cmidrule(lr){4-5}\cmidrule(lr){6-7}
    \textbf{Metric} & Baseline & Pipeline (ours) & Baseline & Pipeline (ours) & Baseline & Pipeline (ours) \\
    \midrule
    hpsv2    & $0.240 \pm 0.024$ & $0.231 \pm 0.063$ & $0.232 \pm 0.030$ & $0.233 \pm 0.024$ & $0.213 \pm 0.036$ & $\mathbf{0.221 \pm 0.027}$ \\
    % ssim   & $0.344 \pm 0.215$ & $\mathbf{0.405 \pm 0.184}^{*}$ & $0.512 \pm 0.171$ & $\mathbf{0.564 \pm 0.143}^{*}$ & $0.453 \pm 0.167$ & $\mathbf{0.498 \pm 0.068}^{*}$ \\
    ms\_ssim & $0.239 \pm 0.210$ & $\mathbf{0.282 \pm 0.179}^{*}$ & $0.382 \pm 0.269$ & $\mathbf{0.422 \pm 0.219}^{*}$ & $0.289 \pm 0.228$ & $\mathbf{0.334 \pm 0.103}^{*}$ \\
    dinov2   & $0.401 \pm 0.282$ & $\mathbf{0.589 \pm 0.274}^{*}$ & $0.645 \pm 0.213$ & $\mathbf{0.781 \pm 0.185}^{*}$ & $0.538 \pm 0.235$ & $\mathbf{0.669 \pm 0.196}^{*}$ \\
    \bottomrule
  \end{tabularx}

  \vspace{2pt}
  {\footnotesize \emph{Note.} $^{*}$ Statistically significant improvement over baseline ($p<0.05$, paired $t$-test).}
\end{table*}

% %%% Add to suplementary
% \begin{table}[h]
% \centering
% \caption{Comparison between \textit{Insert Diffusion} and \textit{Insert Diffusion integrated in our Pipeline}. Values are reported as Mean $\pm$ Std. Statistically significant differences are marked with *.}
% \label{tab:insertDiff}
% \resizebox{\columnwidth}{!}{
% \begin{tabular}{lcc}
% \hline
% \textbf{Metric} & \textbf{Insert Diffusion(Baseline)} & \textbf{Insert Diffusion Pipeline(OURS)} \\
% \hline
% hpsv2\_score       & $0.2279 \pm 0.0271$ & $0.2283 \pm 0.0609$ \\
% ssim               & $0.2887 \pm 0.2107$ & $\mathbf{0.4144 \pm 0.1684}$ * \\
% ms\_ssim           & $0.1803 \pm 0.1728$ & $\mathbf{0.2815 \pm 0.1737}$ * \\
% dinov2   & $0.3062 \pm 0.2431$ & $\mathbf{0.4991 \pm 0.2412}$ * \\
% \hline
% \end{tabular}}
% \end{table}

\begin{figure*}[t]
  % Left-aligned, smaller caption (scoped to this figure)
  \captionsetup{justification=raggedright, singlelinecheck=false, font=small}

  \centering
  \setlength{\tabcolsep}{4pt}
  \renewcommand{\arraystretch}{0.98}

  \begin{tabular}{@{}LCCC@{}}
    \makecell{\textbf{Reference}\\[2pt]\scriptsize (image)} &
    \makecell{\textbf{Nova}\\[1pt]\scriptsize Baseline vs Pipeline} &
    \makecell{\textbf{ICLight}\\[1pt]\scriptsize Baseline vs Pipeline} &
    \makecell{\textbf{Replace Anything}\\[1pt]\scriptsize Baseline vs Pipeline} \\
    \midrule

    % Row 1 (images)
    \RefCell{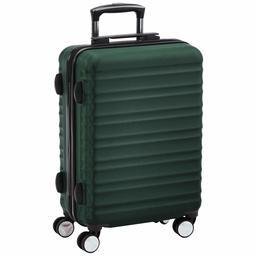} &
    \StackTwo{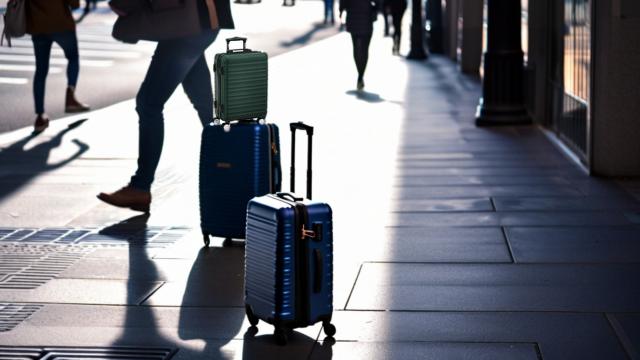}{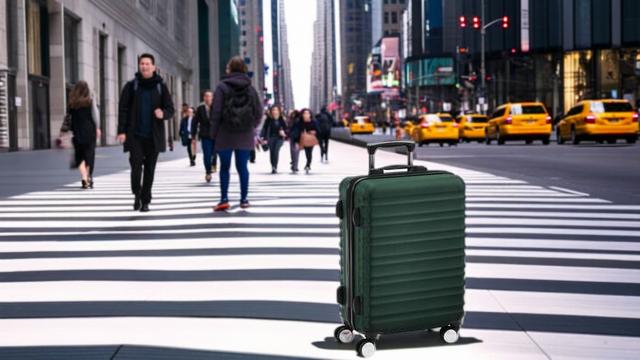} &
    \StackTwo{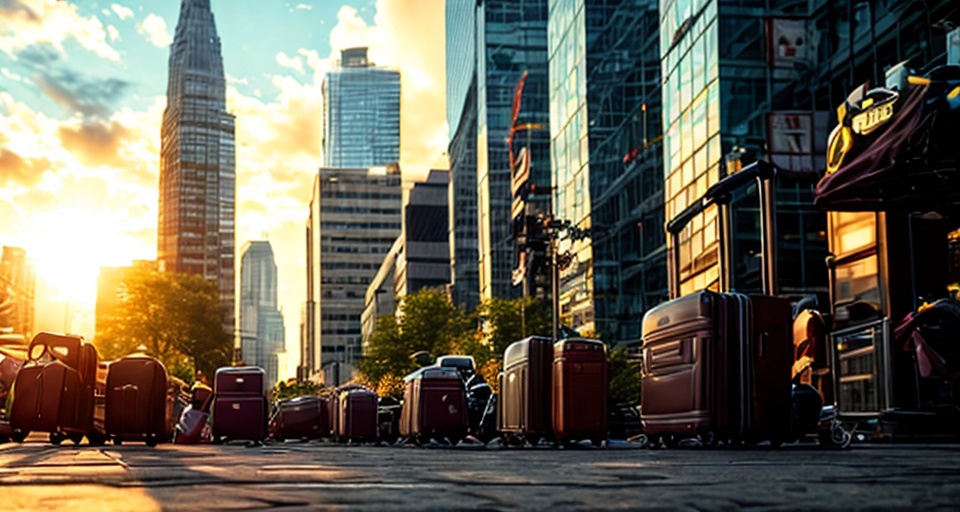}{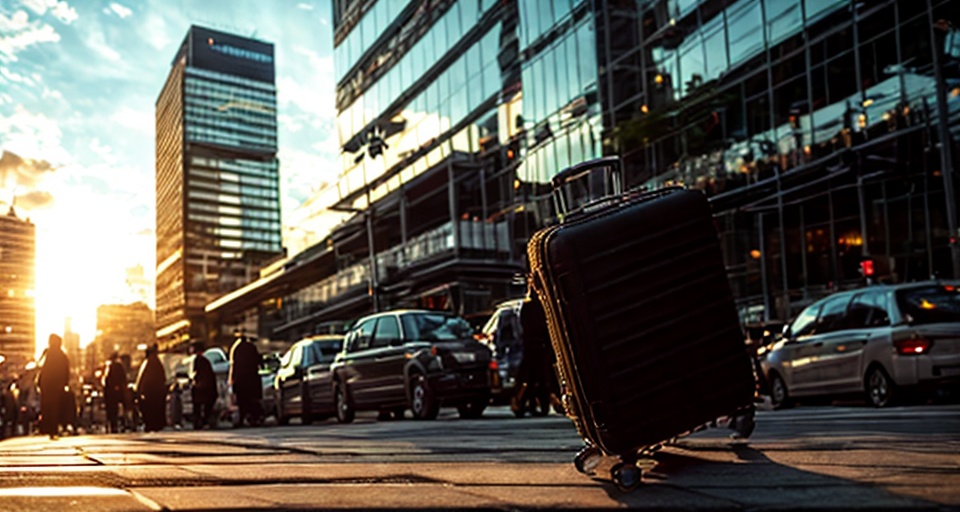} &
    \StackTwo{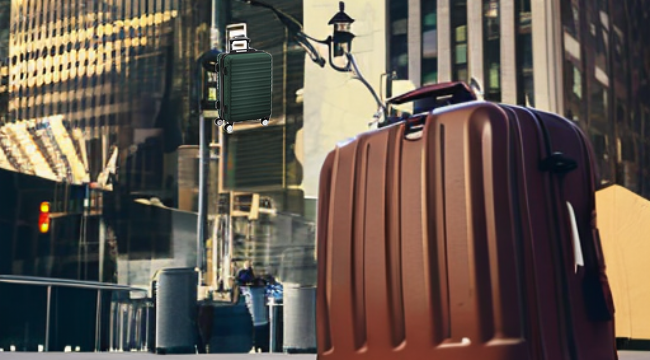}{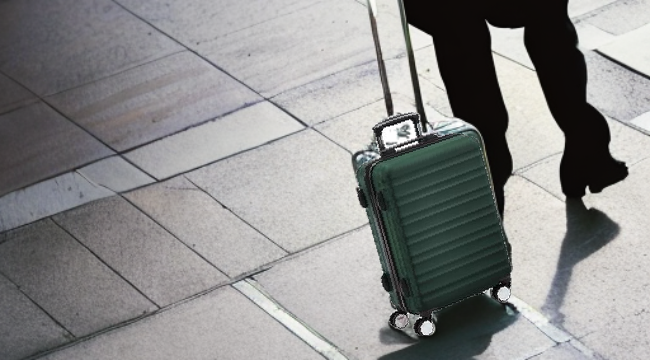} \\
    % Row 1 (prompt spanning all 4 cols)
    \PromptRow{Low angle view of luggage positioned on a busy city sidewalk with towering skyscrapers and bustling pedestrians in the background; dramatic evening light from street lamps casting long shadows.}

    % Row 2 (images)
    \RefCell{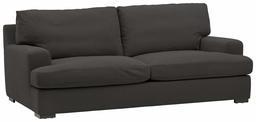} &
    \StackTwo{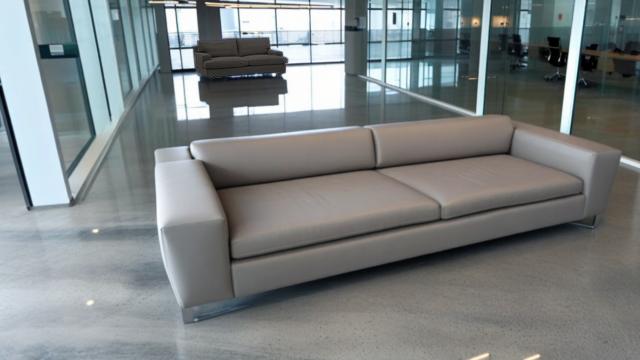}{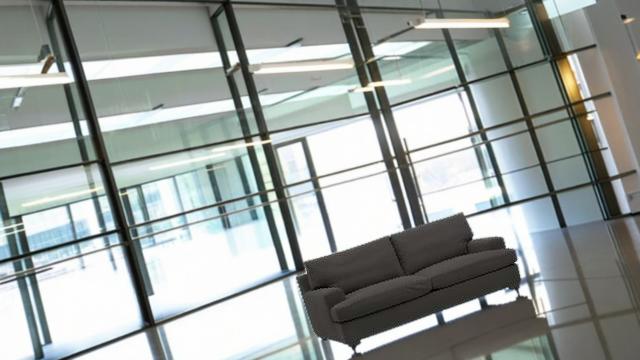} &
    \StackTwo{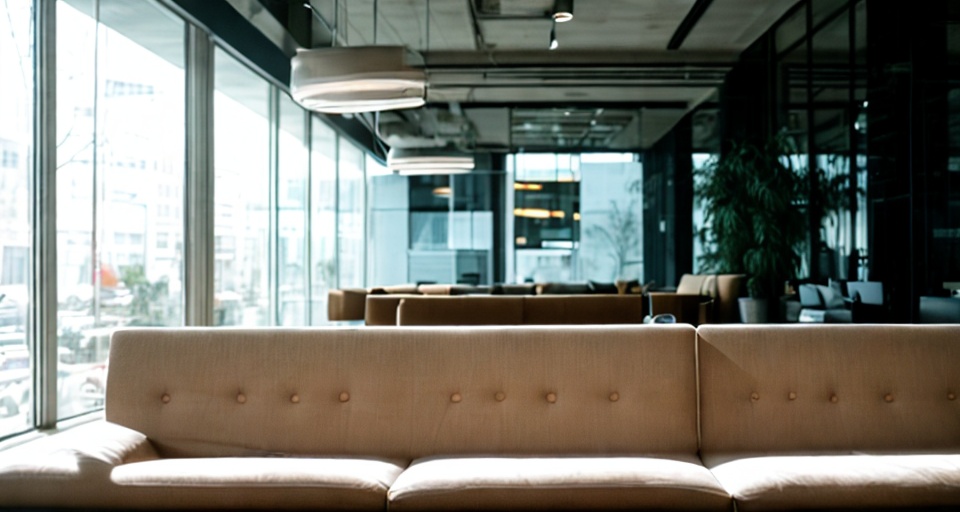}{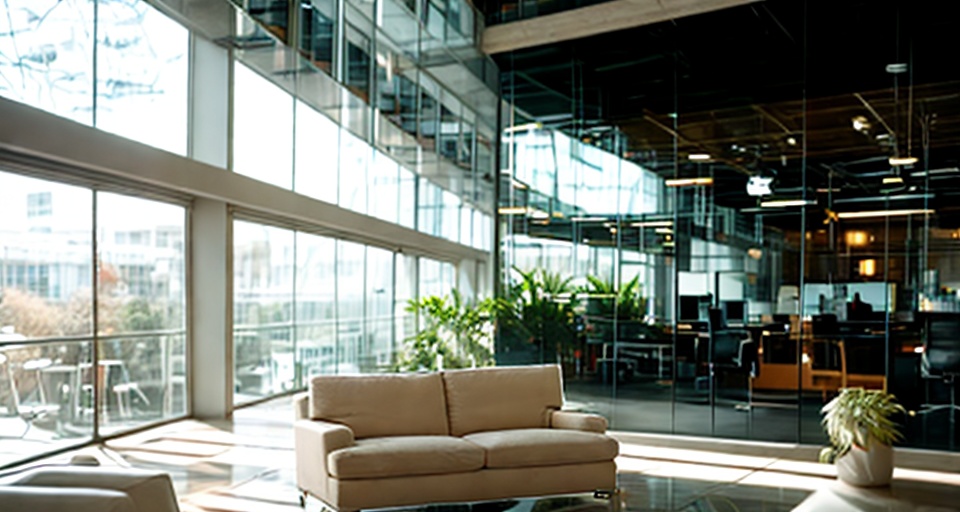} &
    \StackTwo{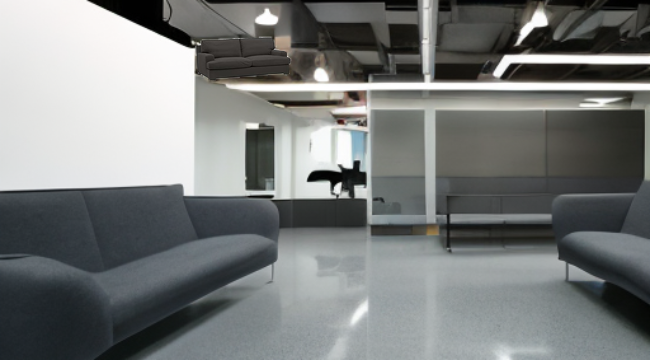}{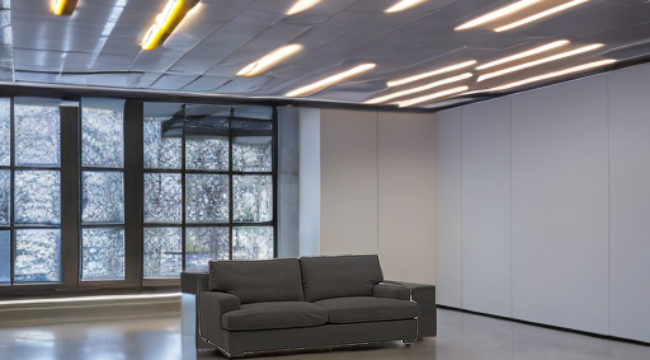} \\
    % Row 2 (prompt spanning all 4 cols)
    \PromptRow{Low-angle view of a sofa in a contemporary office with glass walls and polished concrete floors, under bright fluorescent overhead lighting.}

    % ... rows 3–4 in the same pattern ...
  \end{tabular}

  \caption{Baseline vs. pipeline outputs across three models for multiple product references. Prompts span the full width below each image row.}
  \label{fig:baseline-vs-workflow-grid}
\end{figure*}

\subsection{Qualitative Evaluation}
Evaluating image quality remains a challenging task. While numerous metrics exist, each captures only limited aspects, and none adequately assess whether an object is convincingly integrated into a scene—critical for inpainting and outpainting. Consequently, we rely on qualitative evaluation through a user study. We employ Amazon Ground Truth Labeling Service ~\cite{aws_groundtruth} to compare images generated by the baseline model and our pipeline.

We randomly sample 50 generation pairs per model, shuffle them, and collect responses from five annotators per pair. For each task, the annotators are shown a caption, the original product image, and two generated images depicting the product within the described scene. They are instructed to choose the best generated outcome according to the following criteria (in order of importance):

\textit{Carefully read the provided caption and look at the provided reference image for the product. Compare the two generated images and select the one that matches the below criteria best. The requirements are listed in the order of importance. Make sure to move down to the next requirement only if both of the images have equivalent performance on the current requirement.}
\begin{enumerate}
    \item \textit{Integration of product in the scene}
    \item \textit{Maintaining product look and fidelity}     
    \item \textit{Alignment with caption: The generated image clearly demonstrates the elements described in the caption and their relative positions in the scene.} 
\end{enumerate}

Since our pipeline has a strict quality assessment module, there exist cases where none of the generated images passed our agentic quality assessment. During sampling image pairs, if we sample a rare case like this, we take a note and consider an extra win for the baseline model when the preference rate based on annotations is calculated.

\begin{figure}[t]
    \centering
    \includegraphics[width=\linewidth]{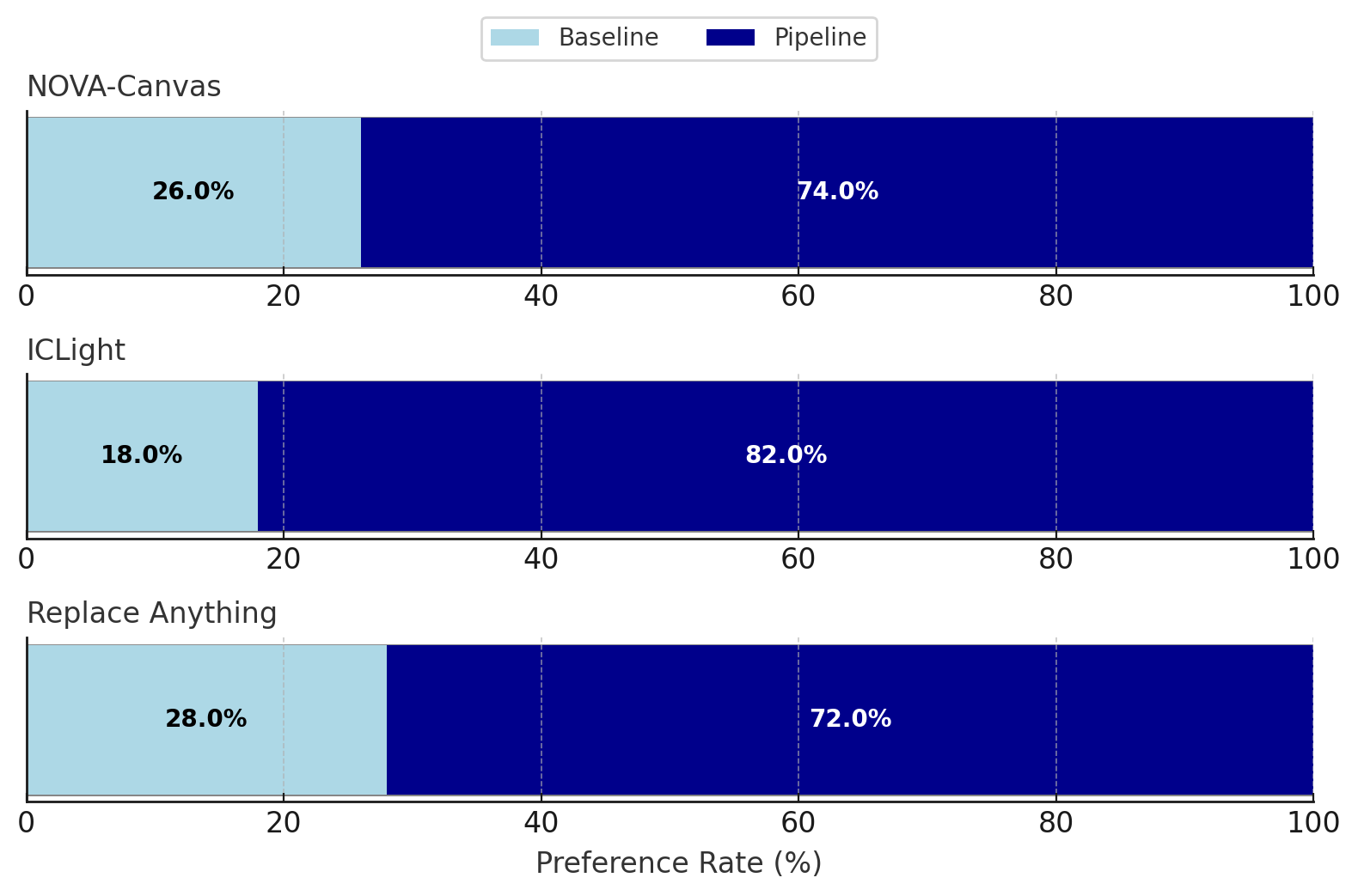} % or .pdf
    \caption{Preference rates for Baseline (light blue) vs. Pipeline (dark blue) across different models.}
    \label{fig:preferences}
\end{figure}

As shown in Figure~\ref{fig:preferences}, the results for three of our test models show a clear preference of the pipeline over the baseline run of the models.

In summary, our numerical metrics show a consistent improvement of the pipeline compared to the base model in maintaining product's integrity and our human evaluation results further show that when choosing a proper base model for the task in hand, pipeline can significantly help the model to generate more quality images.

\subsection{Limitations}
\label{sec:limit}
We identify two primary sources of failure in our pipeline: dataset limitations and base model limitations.  

\subsubsection{Dataset Limitations}
The ABO dataset contains a small number of mis-labeled product images. Although infrequent, these errors can confuse the generation process and lead to nondeterministic outputs (see Supplementary Material, Section 1.2, Figures 7, 8, and 9).  

\subsubsection{Base Model Limitations}
The majority of failure cases stem from the underlying generative models. First, such models often hallucinate by generating spurious element around the masked region. In our set of baseline models we observe this failure case mostly in the outcomes of Replace Anything (see Supplementary Material, Section 1.2, Figure 4). Our placement strategy reduces - but does not eliminate - this effect, and when the base model fails, the pipeline inherits these errors. 

Second, when the reference product appears relatively too small in the scene, models tend to ignore the object entirely. The model in this case might generate a different instance of the product, or omit it altogether. Our current quality module detects the latter case (rule number 2 in Section ~\ref{sec:qualitycontrol}) but does not explicitly check for product similarity to balance cost and latency. Repeated cases of this error type would require the integrating of an object fidelity test into the Quality Control Module (see Supplementary Material, Section 1.2, Figures 5, and 6).  

Finally, individual base models exhibit their own quirks. For example, despite being trained to preserve object features, our experiments reveal pixel-level changes to products when using ICLight as the image generator. Particularly, ICLight often alters product texture or color when adjusting illumination. Since such behaviors are model-specific, we recommend testing the base model on a small subset of data to surface recurring issues and adjusting the quality checks accordingly (see Supplementary Material, Section 1.2, Figure 10).  

%% file: sec/5_conclusion.tex
\section{Conclusion and Future Work}
In this paper, we introduce a modular pipeline for scalable generation of marketing images. Although tailored to marketing, the design is highly customizable and can be adapted to a wide range of content-creation tasks that follow visual guidelines. By automating the iterative generation process and enforcing quality through guardrail checks, our approach reduces the need for constant human oversight. Instead, human effort is focused where it is most valuable—making final selections from images that already satisfy predefined criteria. While our current system focuses primarily on maintaining product integrity and visual composition, brands often require adherence to strict and diverse guidelines such as color palettes, copy style, and layout constraints. Future work includes extending the pipeline to automatically incorporate these brand-specific rules, enabling more comprehensive compliance and further streamlining large-scale production workflows. We view this work as a step toward the productionization of image generation, enabling deployment at scale and addressing real industry needs.

%% file: sec/6_supplementary.tex
\section{Supplementary Material}
\subsection{Additional success examples from the Pipeline}

\begin{figure*}[ht]
% \begin{strip}
  \centering
  \renewcommand{\arraystretch}{1.0}
  \begin{tabular}{@{}JKK@{}} % <-- contiguous, no spaces
    \makecell{\textbf{Reference}\\[2pt]\scriptsize (image + prompt)} &
    \makecell{\textbf{Baseline}} &
    \makecell{\textbf{{Pipeline}}} \\
    \hline
    \SupRow{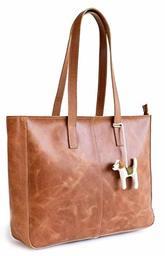}{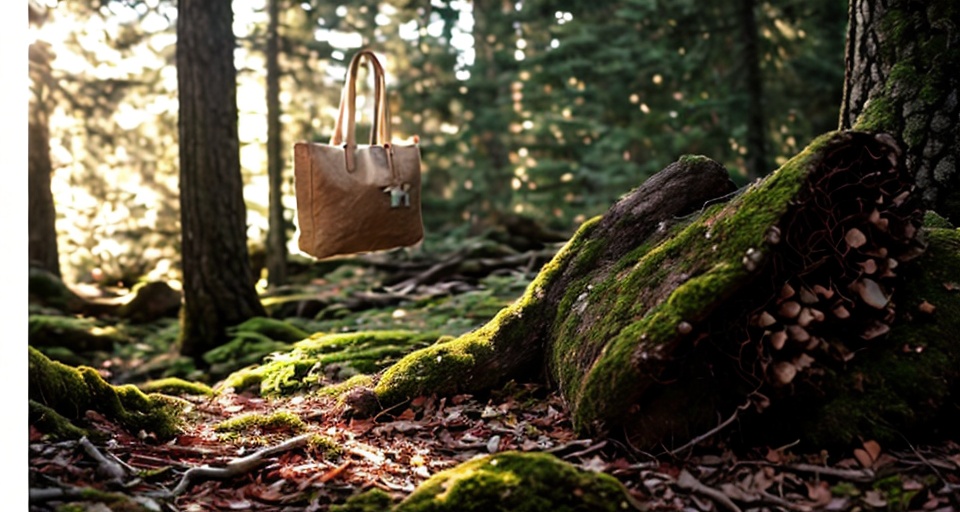}{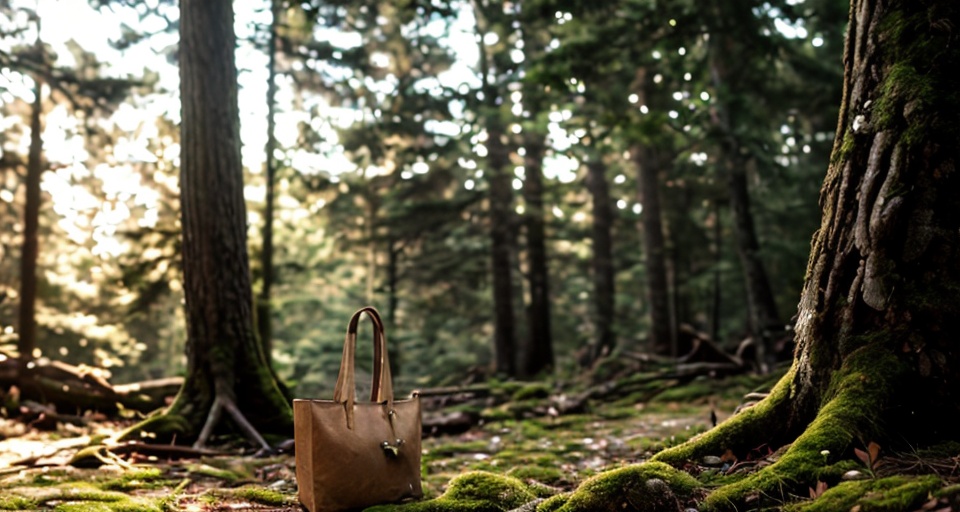}{Wide-angle view of a bag positioned on a moss-covered fallen log in a dense forest clearing surrounded by towering pine trees and scattered wildflowers, soft diffused natural light filtering through the canopy above.}
    
    \SupRow{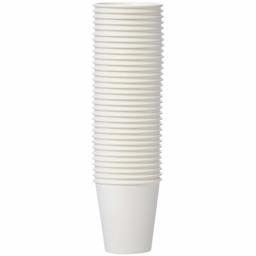}{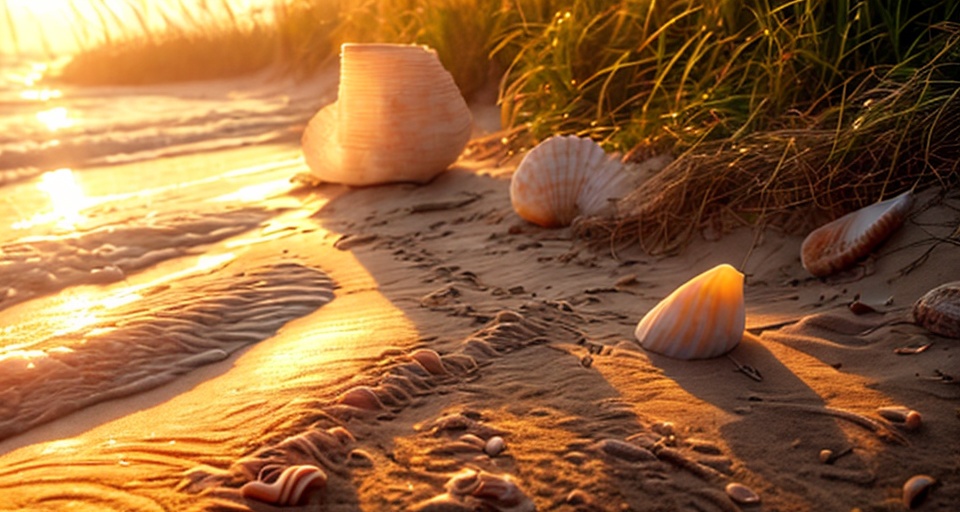}{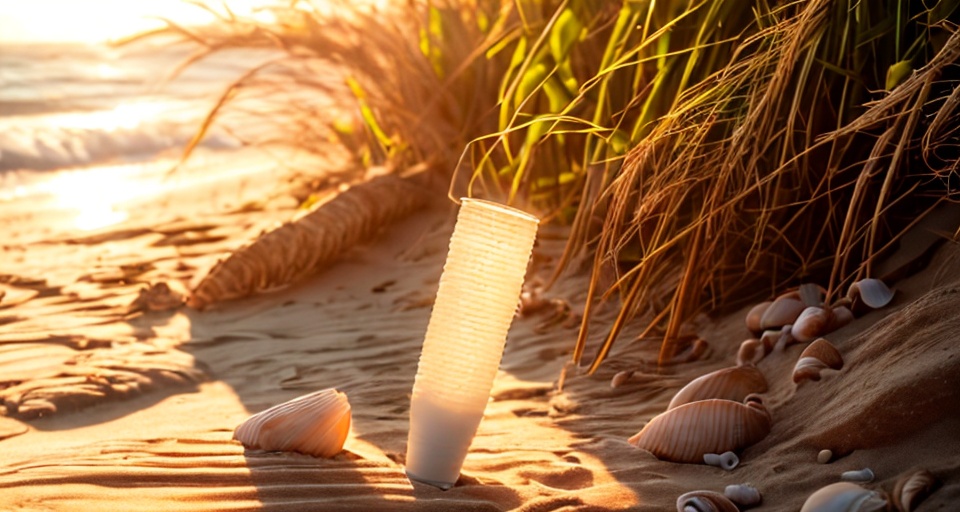}{Low-angle perspective of a drinking cup sitting on sandy beach dunes with ocean waves and seashells visible in the distance, golden hour sunlight casting long dramatic shadows across the scene.}
    
    \SupRow{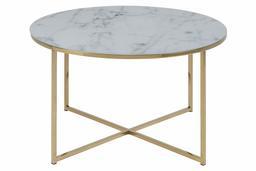}{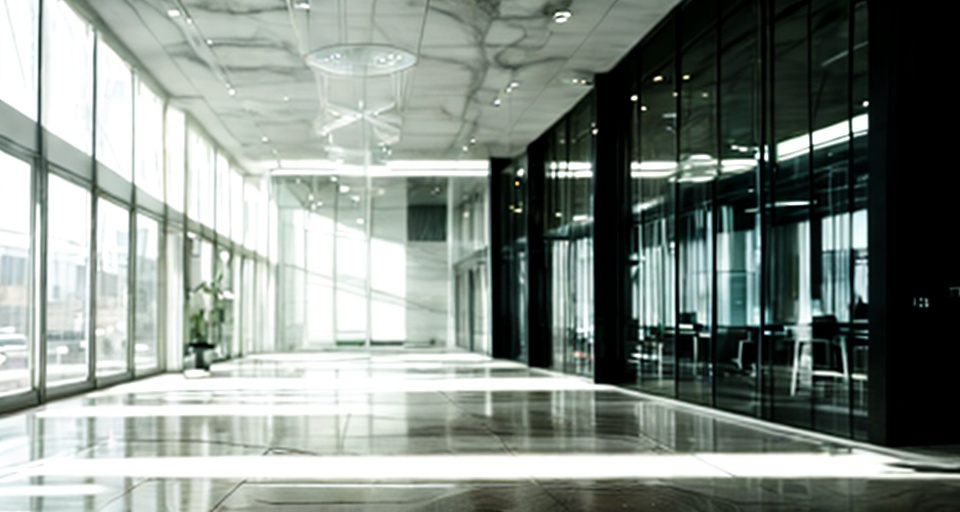}{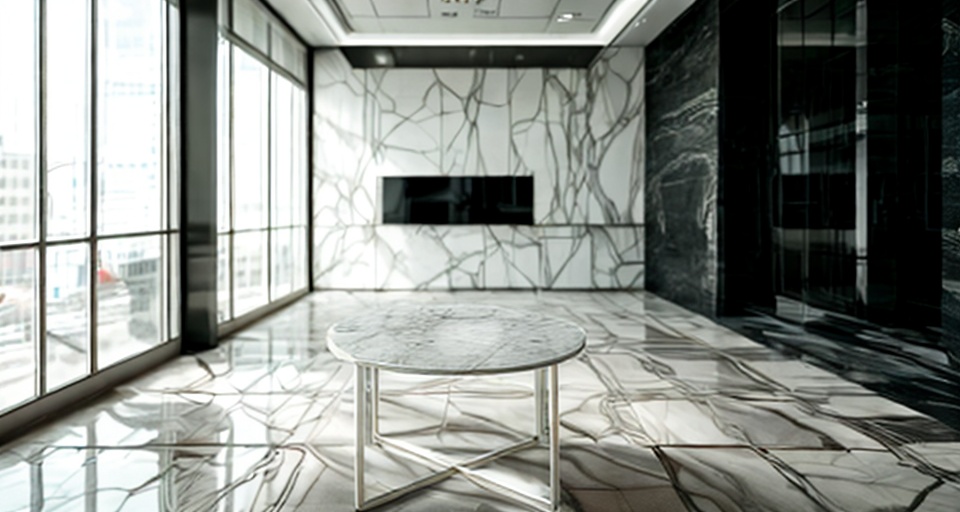}{Medium shot of a table positioned in a sleek modern office space with glass walls, minimalist decor, and polished marble floors, illuminated by clean white fluorescent lighting and natural daylight, professional photo}

  \end{tabular}
  \caption{Additional success cases where pipeline improves generated image output quality for ICLight. Left: reference. Middle/Right: generated images from baseline and pipeline}
  \label{fig:goodimages_iclight}
\end{figure*}
% \end{strip}

\begin{figure*}[!t]
  \centering
  \renewcommand{\arraystretch}{1.0}
  \begin{tabular}{@{}JKK@{}} % <-- contiguous, no spaces
    \makecell{\textbf{Reference}\\[2pt]\scriptsize (image + prompt)} &
    \makecell{\textbf{Baseline}} &
    \makecell{\textbf{{Pipeline}}} \\
    \hline
    \SupRow{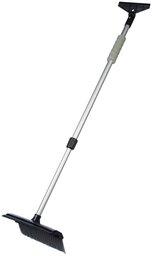}{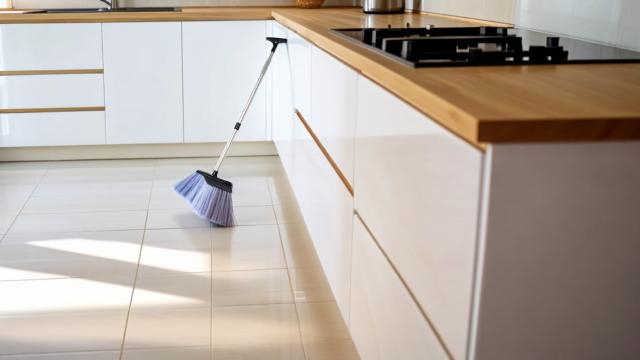}{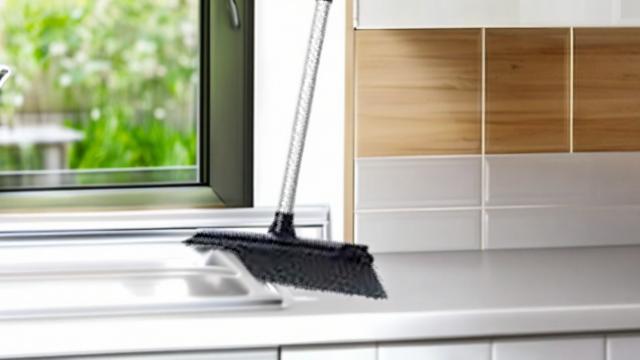}{Close-up shot of a broom leaning against white kitchen cabinets in a modern home interior, ceramic tiles reflecting soft morning sunlight streaming through nearby windows, photo}
    
    \SupRow{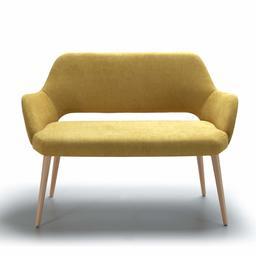}{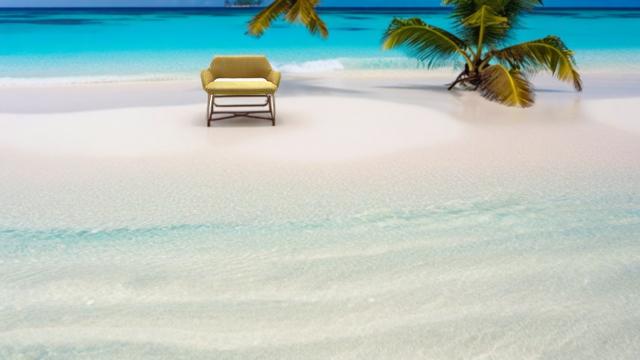}{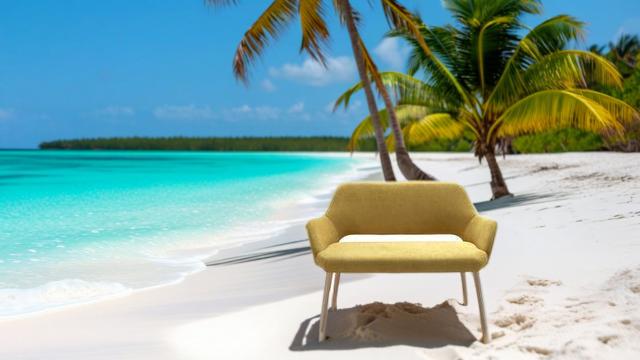}{Medium shot of a chair sitting on pristine white sand at a tropical beach with crystal clear turquoise water, swaying palm trees in the background, and bright midday sunlight creating sharp contrasts and vibrant colors in this paradisiacal coastal scene.}
    
    \SupRow{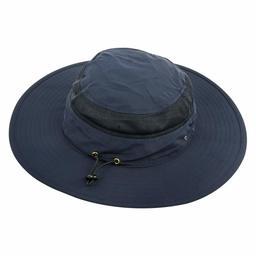}{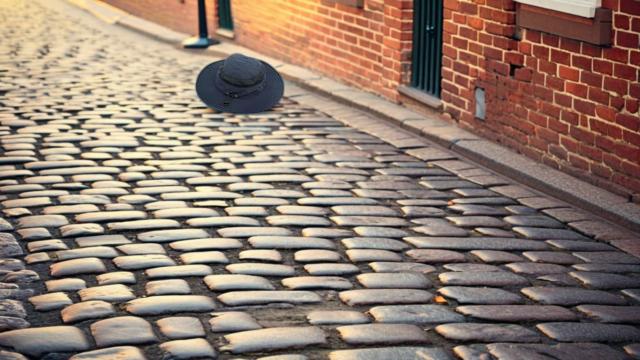}{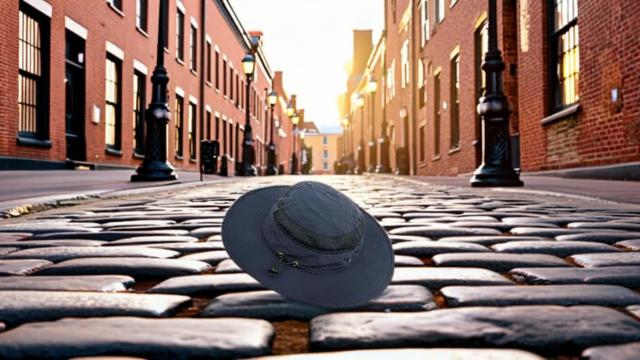}{Close-up shot of a hat resting on weathered cobblestones in a bustling city street with vintage brick buildings and iron lampposts in the background, warm golden hour sunlight casting long shadows across the urban scene, professional street photography style}
    
    \SupRow{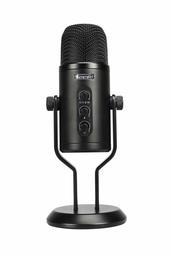}{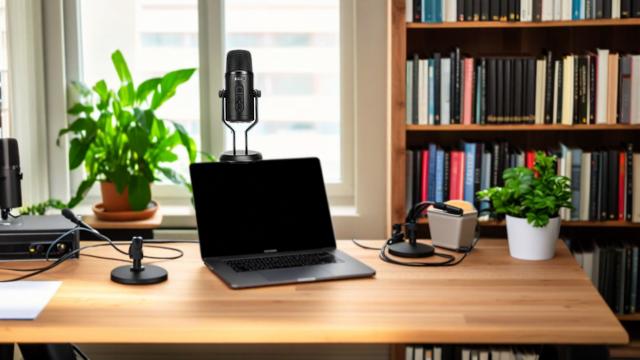}{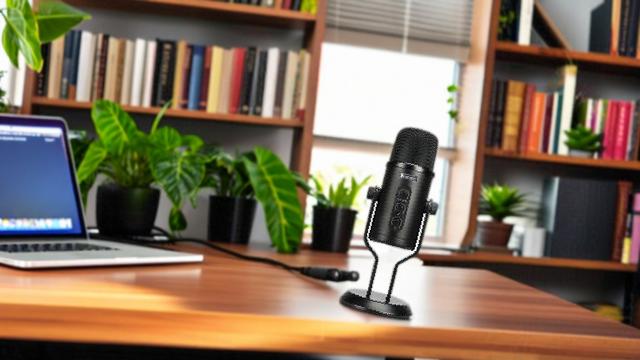}{Medium shot of a microphone positioned on a wooden desk in a cozy home office setup with bookshelves, houseplants, and a laptop computer nearby, soft natural daylight streaming through a nearby window creating gentle shadows and a warm, inviting atmosphere for podcasting.}

  \end{tabular}
  \caption{Additional success cases where pipeline improves generated image output quality for Nova Canvas. Left: reference. Middle/Right: generated images from baseline and pipeline}
  \label{fig:goodimages_nova}
\end{figure*}

\begin{figure*}[!t]
  \centering
  \renewcommand{\arraystretch}{1.0}
  \begin{tabular}{@{}JKK@{}} % <-- contiguous, no spaces
    \makecell{\textbf{Reference}\\[2pt]\scriptsize (image + prompt)} &
    \makecell{\textbf{Baseline}} &
    \makecell{\textbf{{Pipeline}}} \\
    \hline
    \SupRow{supplementary_figures/goodimages/product/bag.jpg}{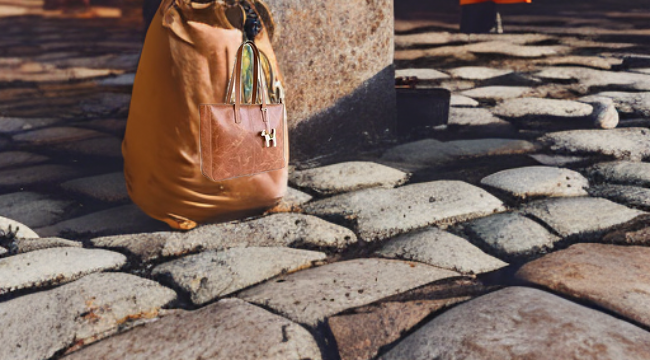}{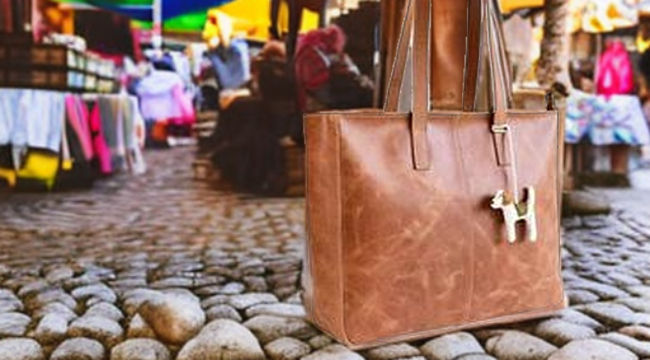}{Close-up shot of a bag resting on weathered cobblestones in a bustling European marketplace with vendors and colorful awnings creating a vibrant backdrop, warm golden hour sunlight casting long shadows across the ancient stone surface."}
    \SupRow{supplementary_figures/goodimages/product/hat.jpg}{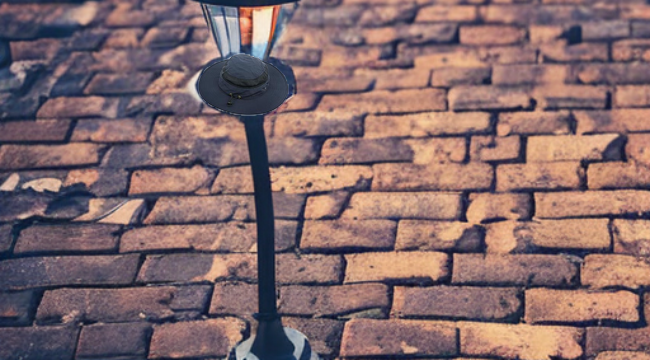}{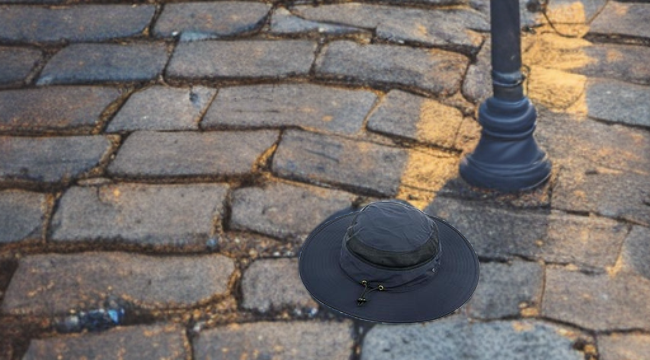}{Close-up shot of a hat resting on weathered cobblestones in a bustling city street with vintage brick buildings and iron lampposts in the background, warm golden hour sunlight casting long shadows across the urban scene, professional street photography style}
    \SupRow{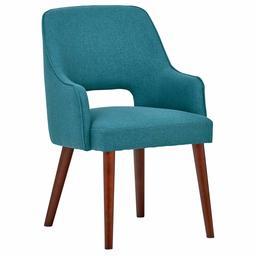}{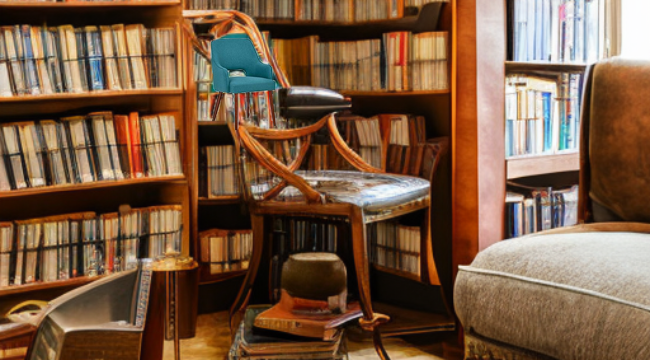}{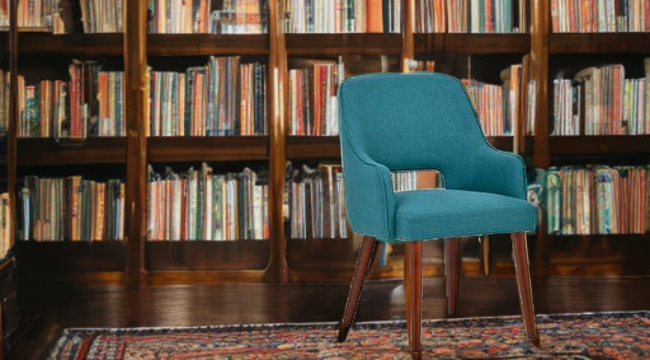}{Close-up angled shot of a chair positioned in a cozy library corner surrounded by towering bookshelves filled with leather-bound volumes, a vintage Persian rug beneath, and warm amber lighting from antique brass lamps creating an intimate scholarly atmosphere.}

  \end{tabular}
  \caption{Additional success cases where pipeline improves generated image output quality for Replace Anything. Left: reference. Middle/Right: generated images from baseline and pipeline}
  \label{fig:goodimages_replaceanything}
\end{figure*}

\clearpage
\subsection{Failure Cases}
\begin{figure*}[ht]
  \centering
  \renewcommand{\arraystretch}{1.0}
  \begin{tabular}{@{}JKK@{}} % <-- contiguous, no spaces
    \makecell{\textbf{Reference}\\[2pt]\scriptsize (image + prompt)} &
    \makecell{\textbf{Baseline}} &
    \makecell{\textbf{{Pipeline}}} \\
    \hline
    \SupRow{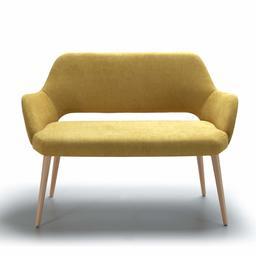}{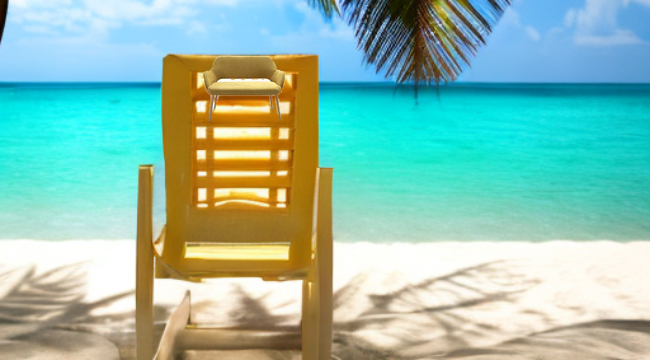}{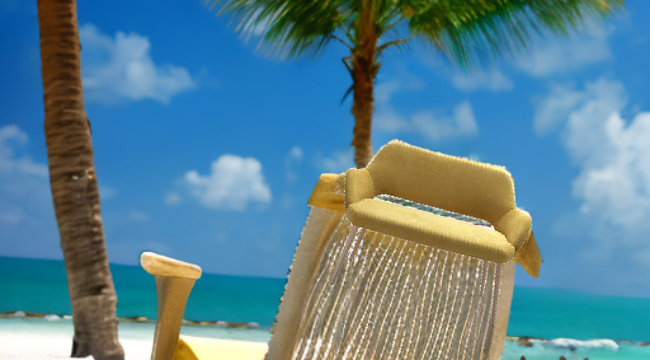}{Medium shot of a chair sitting on pristine white sand at a tropical beach with crystal clear turquoise water, swaying palm trees in the background, and bright midday sunlight creating sharp contrasts and vibrant colors in this paradisiacal coastal scene}
    \SupRow{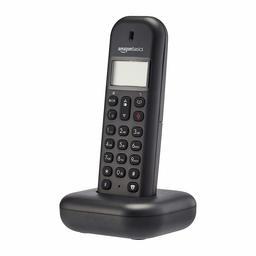}{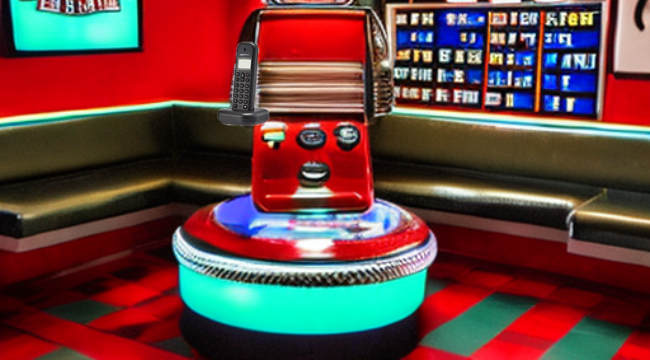}{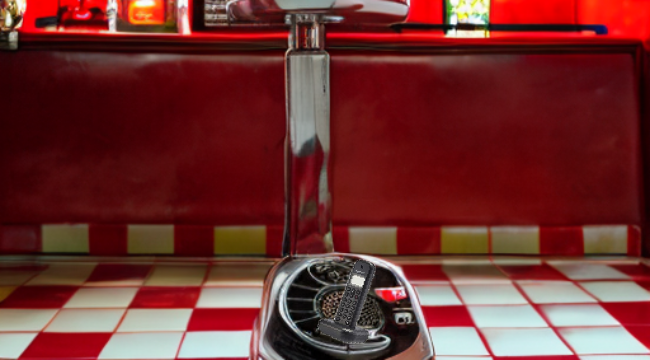}{Wide-angle view of a landline phone positioned on a red vinyl booth table inside a retro American diner with checkered floor tiles, chrome fixtures, and neon signs glowing in the background, bright fluorescent lighting casting even illumination throughout the space.}
    \SupRow{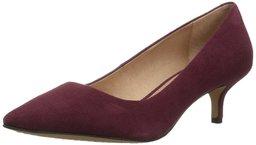}{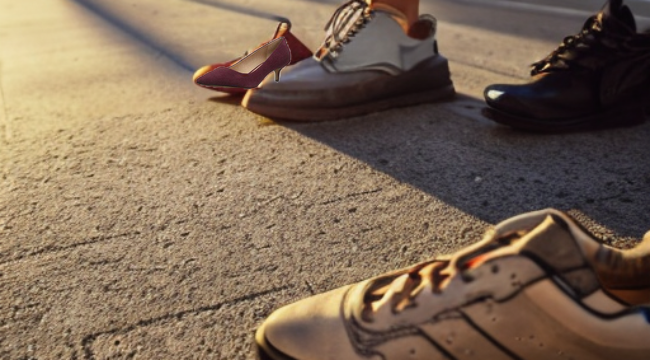}{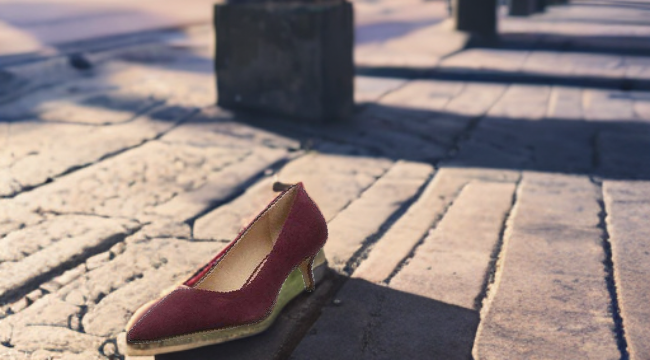}{Close-up shot of shoes resting on weathered concrete pavement in a bustling urban street with blurred city buildings and storefronts in the background, warm golden hour sunlight casting long shadows across the sidewalk, photo}

  \end{tabular}
  \caption{Failure cases for Replace Anything - hallucination. Left: reference. Middle/Right: generated images from baseline and pipeline}
  \label{fig:replaceanything_hallucination}
\end{figure*}

\begin{figure*}[!htbp]
  \centering
  \renewcommand{\arraystretch}{1.0}
  \begin{tabular}{@{}JKK@{}} % <-- contiguous, no spaces
    \makecell{\textbf{Reference}\\[2pt]\scriptsize (image + prompt)} &
    \makecell{\textbf{Baseline}} &
    \makecell{\textbf{{Pipeline}}} \\
    \hline
    \SupRow{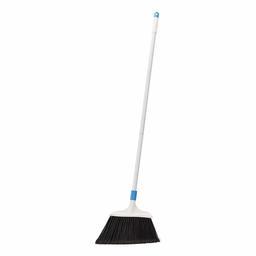}{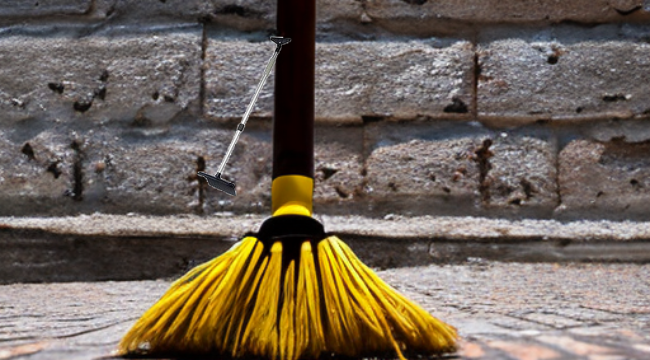}{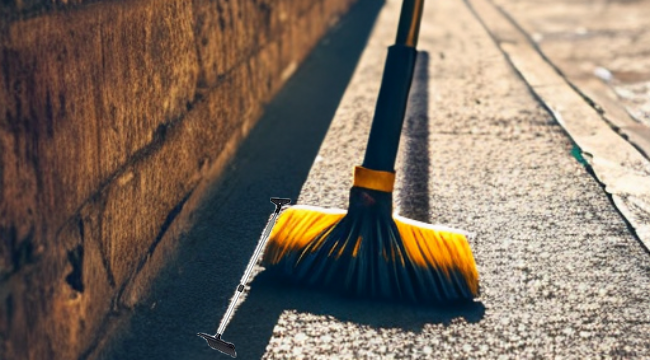}{Low-angle perspective of a broom resting against weathered brick wall in narrow urban alleyway, dramatic evening light from street lamp casting long shadows across wet pavement, photo}
    \SupRow{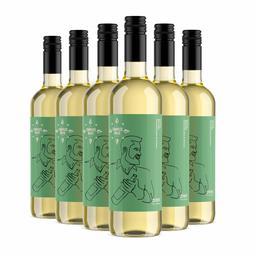}{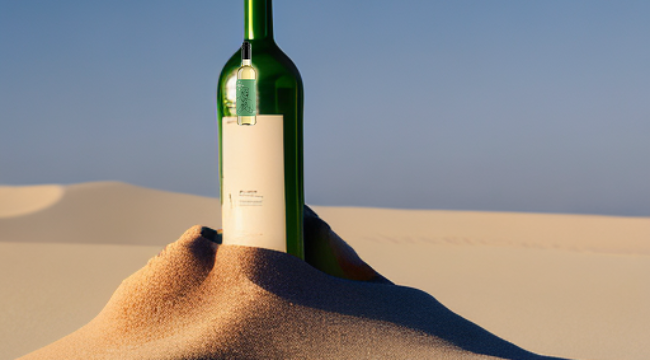}{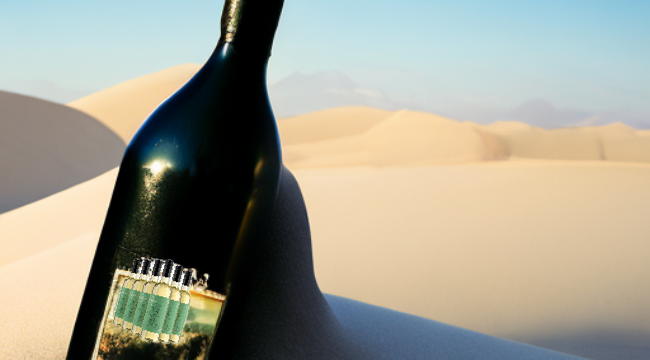}{Low-angle photograph of wine bottle partially buried in white sand dunes with desert landscape and distant mountains visible on the horizon under bright midday sun creating dramatic shadows and harsh contrasts.}

  \end{tabular}
  \caption{Failure cases for Replace Anything - Object Ignored. Left: reference. Middle/Right: generated images from baseline and pipeline. In the above cases the model generated another instance of the object.}
  \label{fig:replaceanything_objectignored}
\end{figure*}

\begin{figure*}[!htbp]
  \centering
  \renewcommand{\arraystretch}{1.0}
  \begin{tabular}{@{}JKK@{}} % <-- contiguous, no spaces
    \makecell{\textbf{Reference}\\[2pt]\scriptsize (image + prompt)} &
    \makecell{\textbf{Baseline}} &
    \makecell{\textbf{{Pipeline}}} \\
    \hline
    \SupRow{supplementary_figures/objectignored/product/broom.jpg}{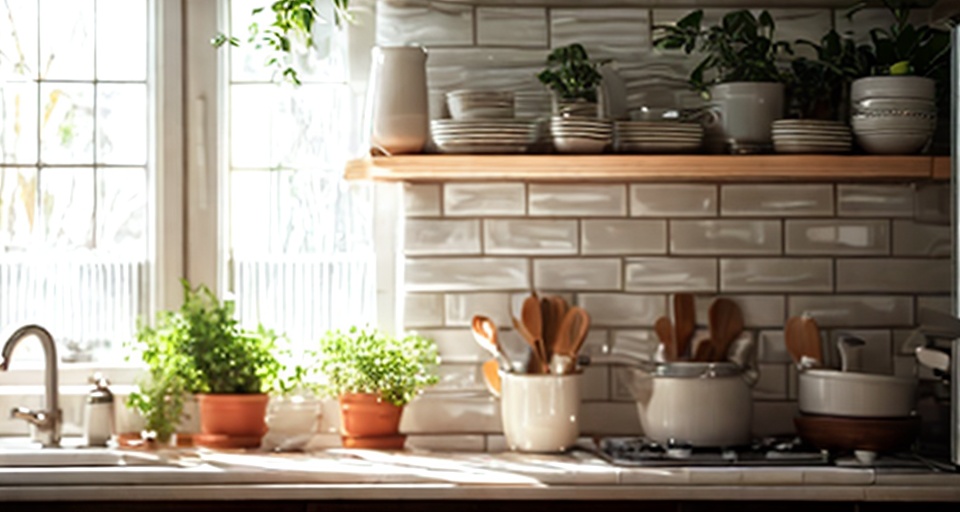}{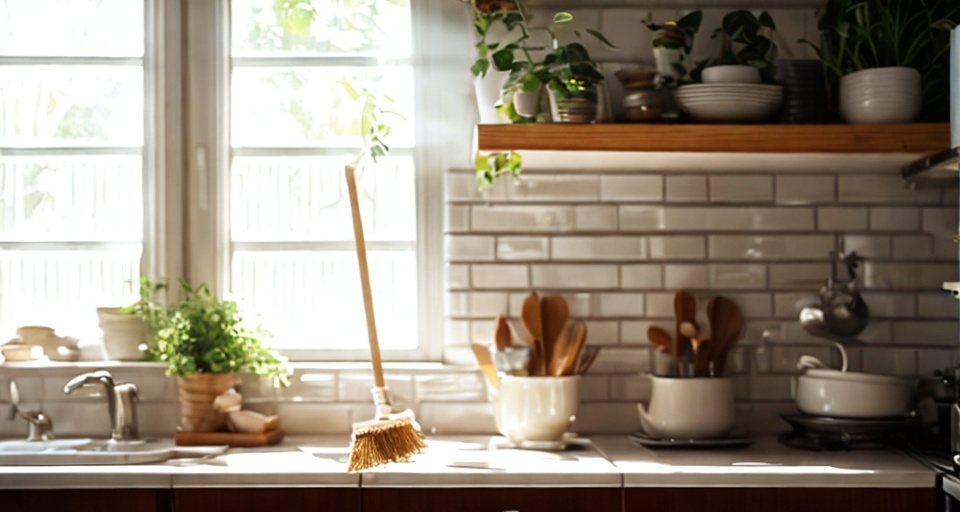}{Close-up shot of a broom leaning against white kitchen cabinets in a modern home interior, ceramic tiles reflecting soft morning sunlight streaming through nearby windows, photo}
    \SupRow{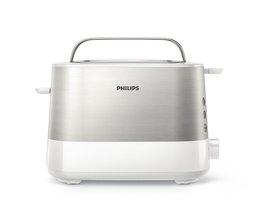}{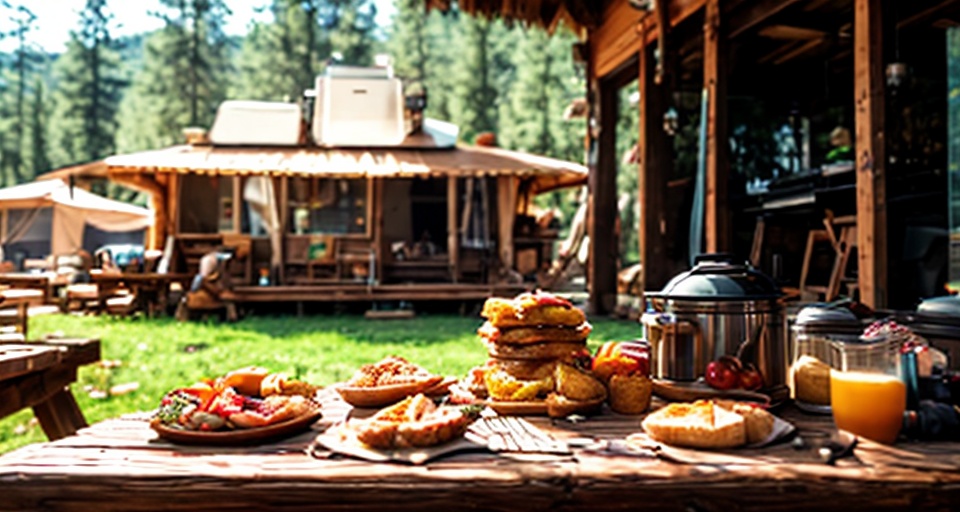}{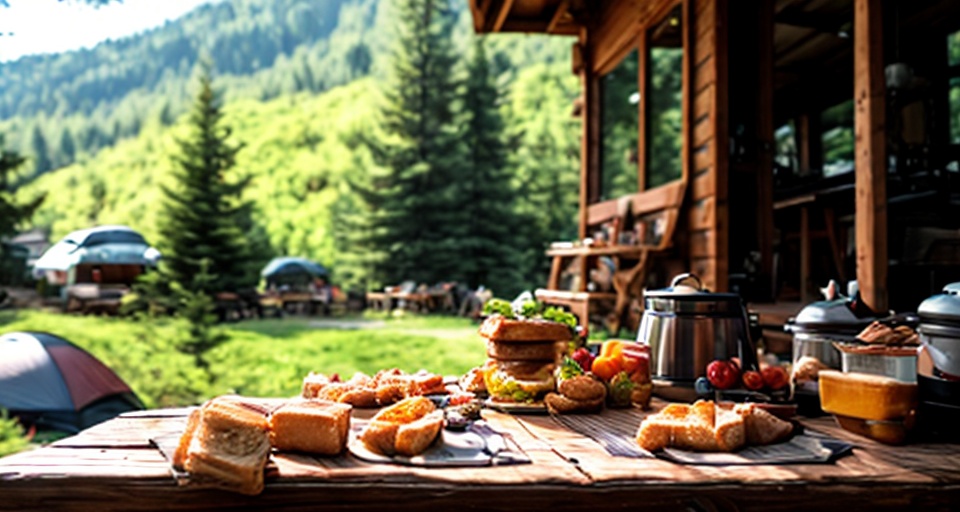}{Wide-angle view of a toaster positioned on a wooden picnic table in a mountain campsite with pine trees and tents visible in the background, bright natural daylight illuminating the outdoor breakfast setup.}

  \end{tabular}
  \caption{Failure cases for ICLight - Object Ignored. Left: reference. Middle/Right: generated images from baseline and pipeline. In the first case the baseline completely ignored the reference image while the pipeline version generated another instance. In the second case, the toaster is floating in the air in the baseline image and is completely ignored in the pipeline generated image.}
  \label{fig:iclight_objectignored}
\end{figure*}

\begin{figure*}[!t]
   \centering
   \begin{subfigure}{\textwidth}
    \centering
% \FloatBarrier
% \begin{figure*}[ht]
%   \centering
  \renewcommand{\arraystretch}{1.0}
  \begin{tabular}{@{}JKK@{}} % <-- contiguous, no spaces
    \makecell{\textbf{Reference}\\[2pt]\scriptsize (image + prompt)} &
    \makecell{\textbf{Baseline}} &
    \makecell{\textbf{{Pipeline}}} \\
    \hline
    \SupRow{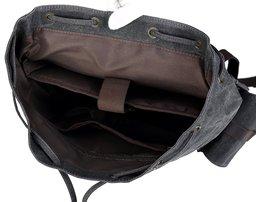}{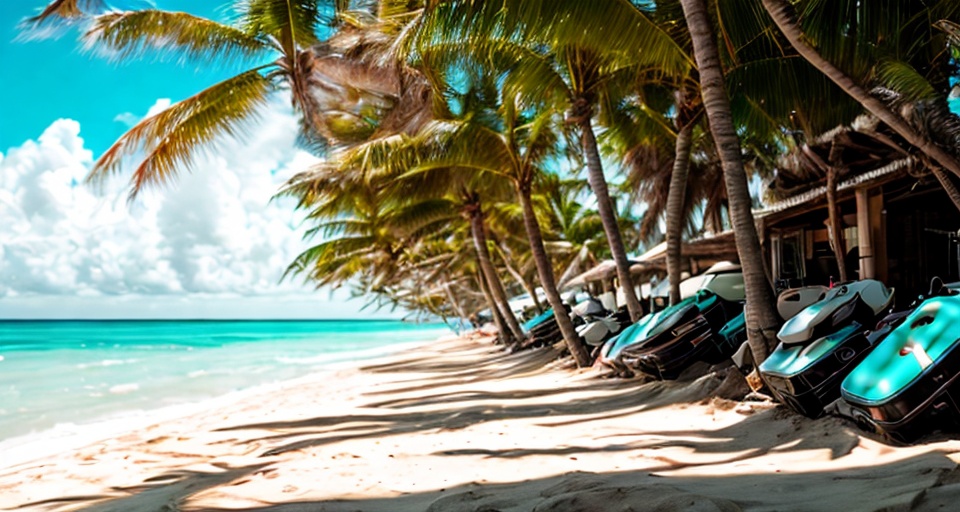}{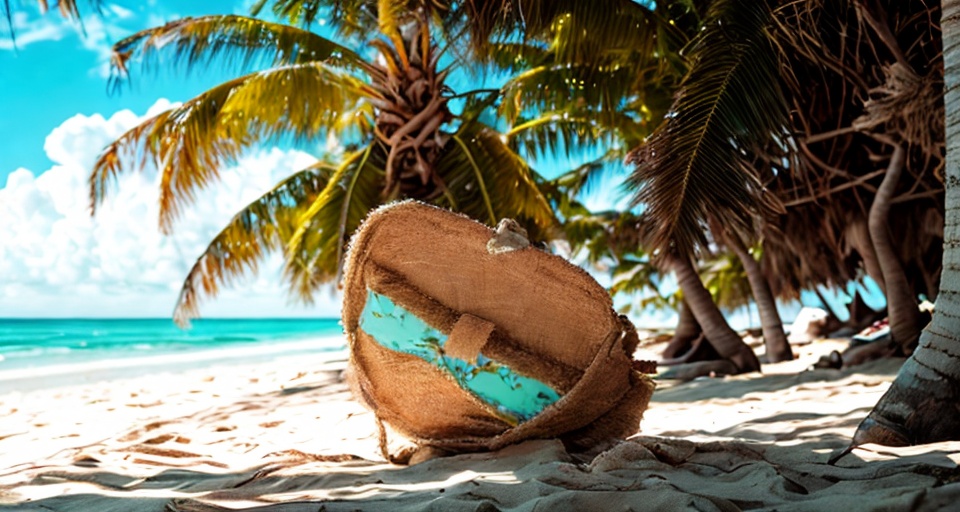}{Medium shot of luggage resting on sandy beach with palm trees swaying and turquoise ocean waves in background, bright tropical sunlight creating vibrant colors and gentle shadows on white sand, photo}
    \SupRow{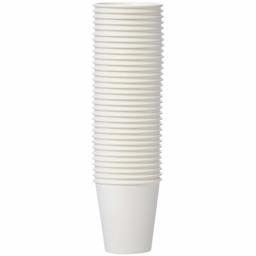}{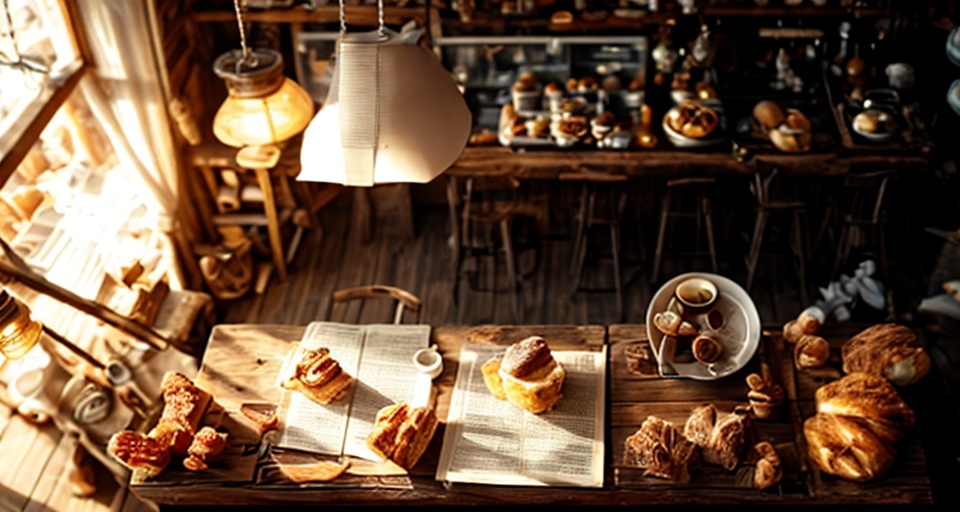}{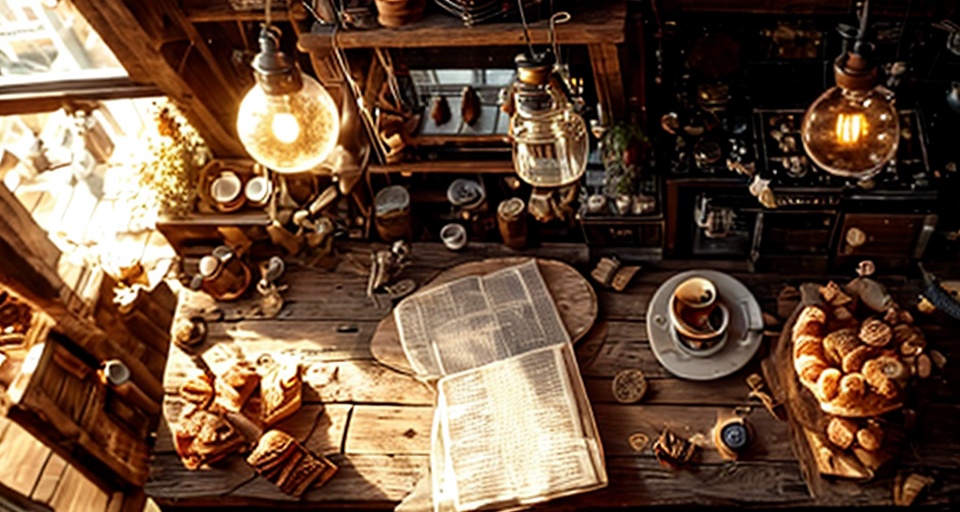}{Overhead shot of a drinking cup placed on a rustic wooden table in a cozy cafe setting with pastries and newspapers scattered around, warm ambient lighting from vintage pendant lamps.}

  \end{tabular}
  \caption{Failure cases for ICLight - Product Mismatch. Left: reference. Middle/Right: generated images from baseline and pipeline. The first example shows a bag labeled as “Luggage” in the ABO dataset. The baseline completely disregards the reference image. In the second row, a stack of cups labeled as “Drinking Cup” confuses the model, leading both the baseline and the pipeline to ignore the provided image reference.}
  \label{fig:iclight_productmismatch}
% \end{figure*}
  \end{subfigure}

 \vspace{1.2em} % vertical space between figures
% \FloatBarrier
% \begin{figure*}[ht]
%   \centering
  % Second figure
  \begin{subfigure}{\textwidth}
    \centering
  \renewcommand{\arraystretch}{1.0}
  \begin{tabular}{@{}JKK@{}} % <-- contiguous, no spaces
    \makecell{\textbf{Reference}\\[2pt]\scriptsize (image + prompt)} &
    \makecell{\textbf{Baseline}} &
    \makecell{\textbf{{Pipeline}}} \\
    \hline
    \SupRow{supplementary_figures/productmismatch/product/bag.jpg}{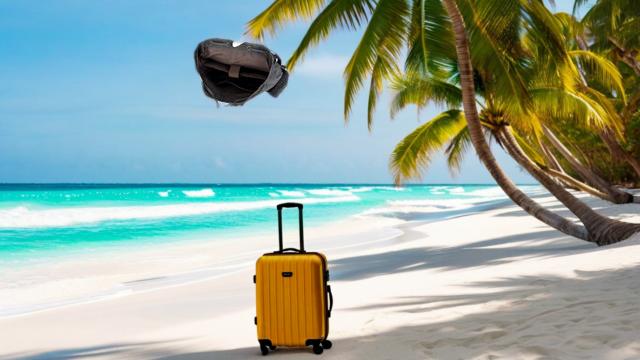}{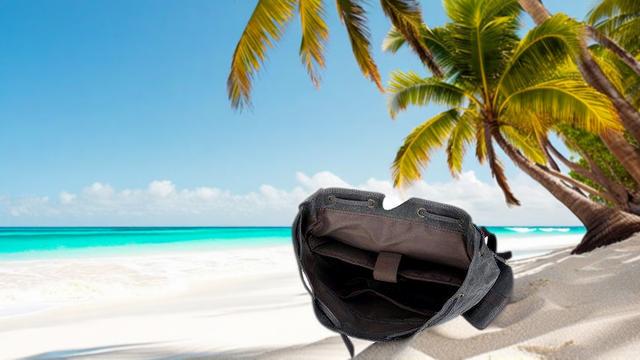}{Medium shot of luggage resting on sandy beach with palm trees swaying and turquoise ocean waves in background, bright tropical sunlight creating vibrant colors and gentle shadows on white sand, photo}
    \SupRow{supplementary_figures/productmismatch/product/cup.jpg}{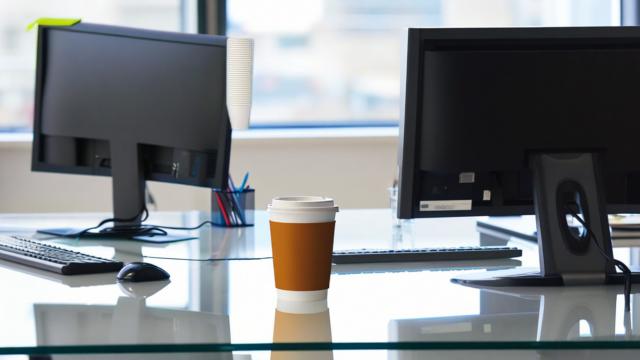}{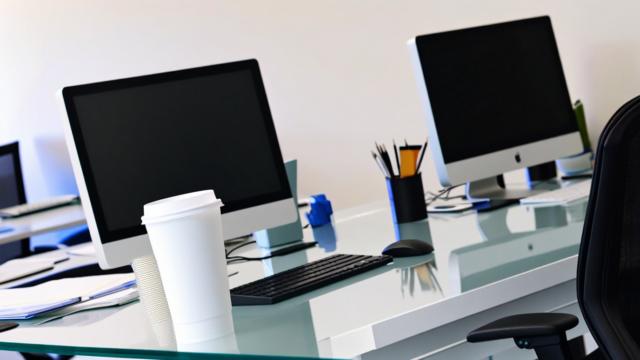}{Eye-level photograph of a drinking cup positioned on a sleek glass desk in a modern office space with computer monitors and office supplies nearby, bright fluorescent lighting creating clean reflections.}

  \end{tabular}
  \caption{Failure cases for Nova Canvas - Product Mismatch. Left: reference. Middle/Right: generated images from baseline and pipeline. In the first row, the bag is labeled as “Luggage” in the ABO dataset. The baseline image shows the model generating a generic piece of luggage, reflecting confusion from the label. In the second row, a stack of reusable cups labeled as “Drinking Cup” results in the model generating coffee cups, capturing only a loose semantic resemblance to the reference image.}
  \label{fig:nova_productmismatch}
% \end{figure*}
  \end{subfigure}

\end{figure*}

\begin{figure*}[ht]
  \centering
  \renewcommand{\arraystretch}{1.0}
  \begin{tabular}{@{}JKK@{}} % <-- contiguous, no spaces
    \makecell{\textbf{Reference}\\[2pt]\scriptsize (image + prompt)} &
    \makecell{\textbf{Baseline}} &
    \makecell{\textbf{{Pipeline}}} \\
    \hline
    \SupRow{supplementary_figures/productmismatch/product/bag.jpg}{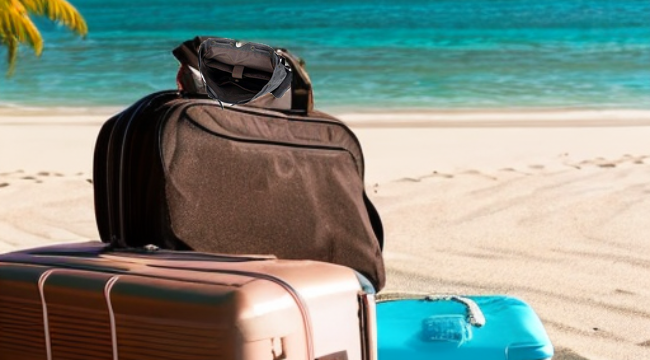}{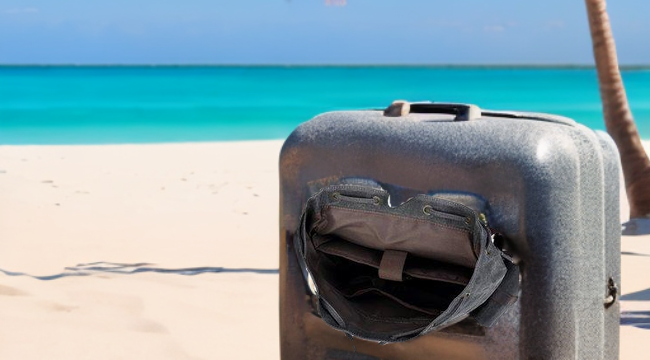}{Medium shot of luggage resting on sandy beach with palm trees swaying and turquoise ocean waves in background, bright tropical sunlight creating vibrant colors and gentle shadows on white sand, photo}
    \SupRow{supplementary_figures/productmismatch/product/cup.jpg}{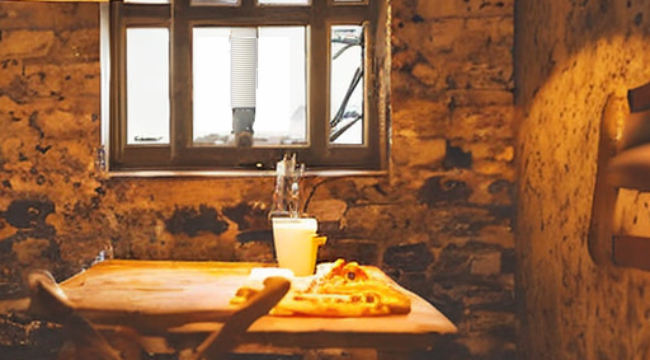}{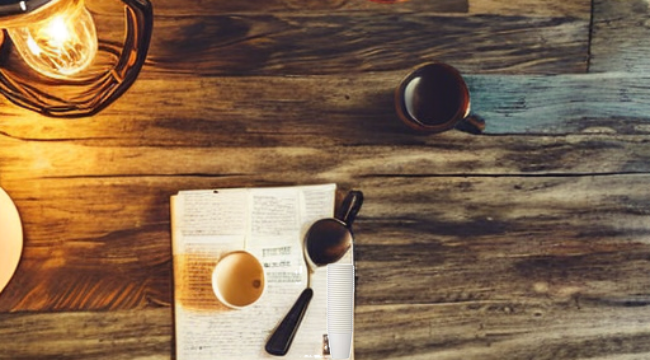}{Overhead shot of a drinking cup placed on a rustic wooden table in a cozy cafe setting with pastries and newspapers scattered around, warm ambient lighting from vintage pendant lamps.}

  \end{tabular}
  \caption{Failure cases for Replace Anything - Product Mismatch. Left: reference. Middle/Right: generated images from baseline and pipeline. In the first row, label confusion causes both methods to generate additional luggage around the reference image. In the second row, while the stack of cups is preserved in the baseline, the model inserts an extra cup on the table, and the pipeline produces a paper cup instead..}
  \label{fig:replaceanything_productmismatch}
\end{figure*}

\begin{figure*}[!t]
  \centering
  \renewcommand{\arraystretch}{1.0}
  \begin{tabular}{@{}JKK@{}} % <-- contiguous, no spaces
    \makecell{\textbf{Reference}\\[2pt]\scriptsize (image + prompt)} &
    \makecell{\textbf{Baseline}} &
    \makecell{\textbf{{Pipeline}}} \\
    \hline
    \SupRow{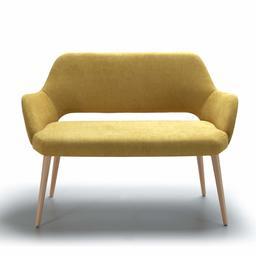}{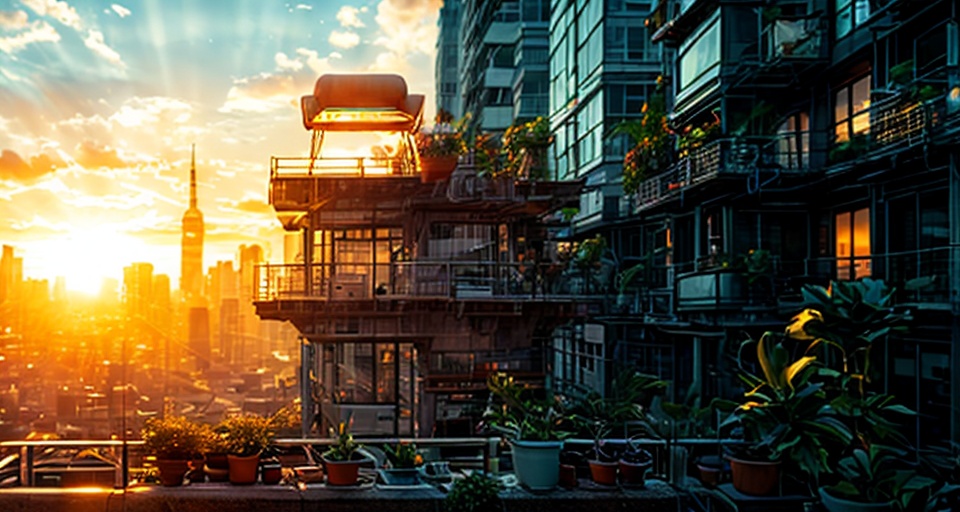}{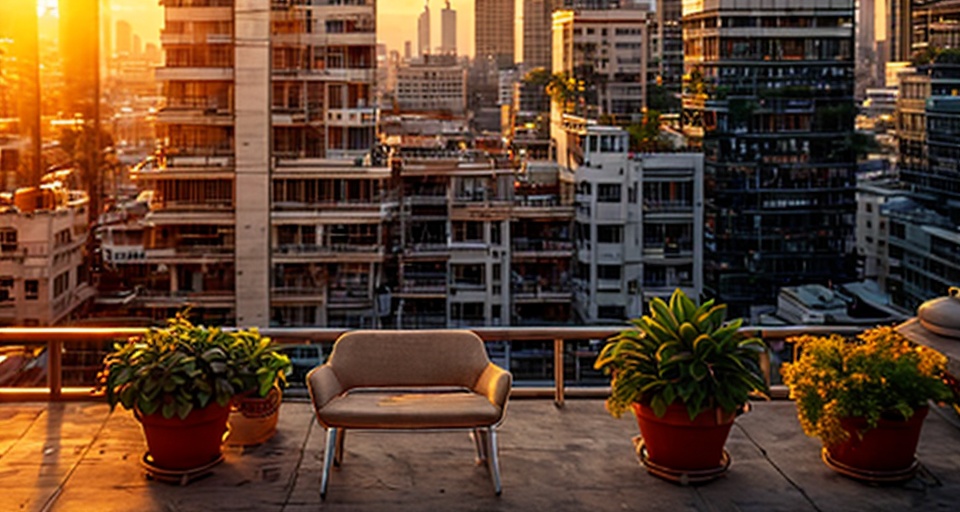}{Bird's eye view of a chair placed on a concrete urban rooftop terrace overlooking a bustling cityscape with skyscrapers, busy streets below, and potted plants around the edges, illuminated by dramatic sunset lighting casting long shadows across the industrial setting.}
    
    \SupRow{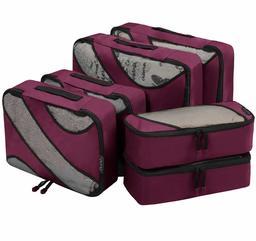}{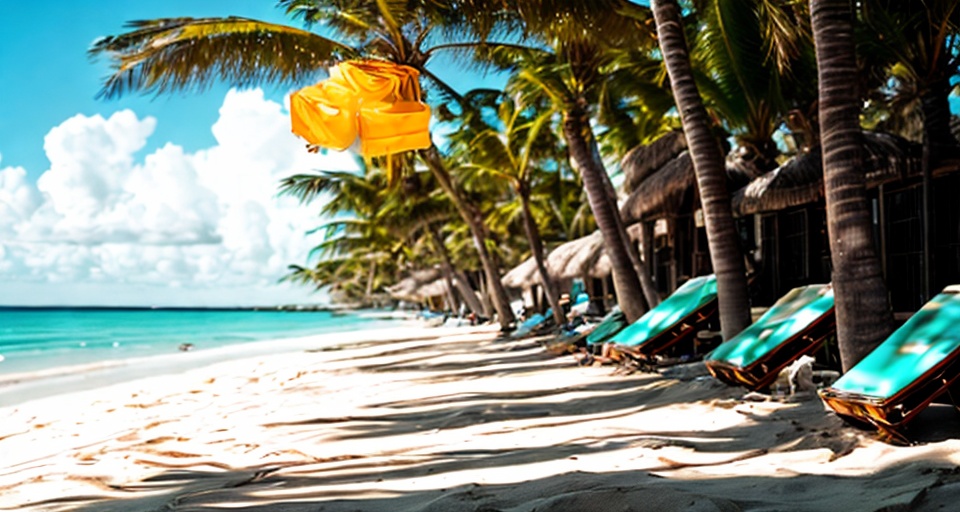}{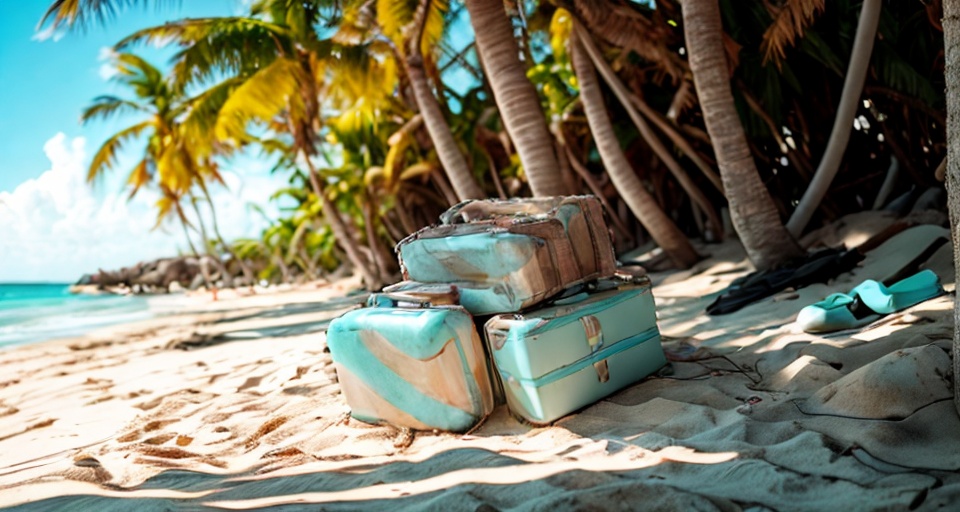}{Medium shot of luggage resting on sandy beach with palm trees swaying and turquoise ocean waves in background, bright tropical sunlight creating vibrant colors and gentle shadows on white sand, photo}

    \SupRow{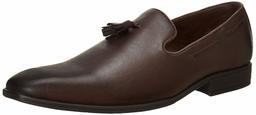}{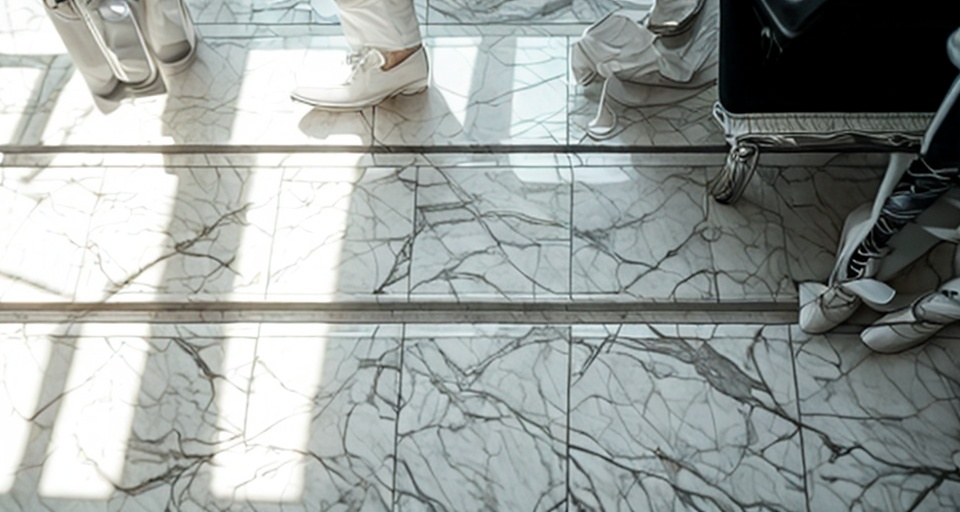}{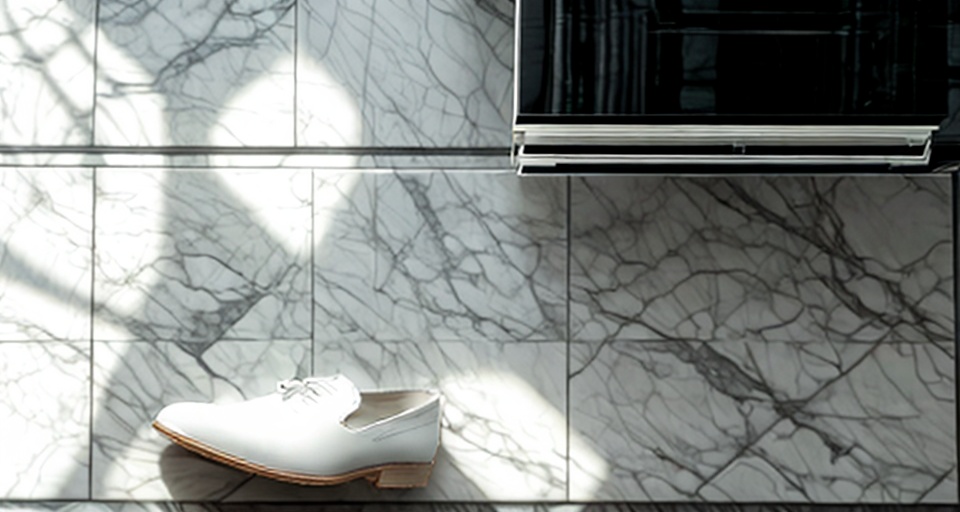}{Overhead shot of shoes placed on polished white marble floor in a minimalist modern interior with clean geometric lines, sleek furniture, and bright even lighting from recessed ceiling fixtures, photo}

  \end{tabular}
  \caption{Failure cases for ICLight - Color Change. Left: reference. Middle/Right: generated images from baseline and pipeline. }
  \label{fig:iclight_colorchange}
\end{figure*}

Figure ~\ref{fig:replaceanything_hallucination} illustrates Replace Anything’s tendency to hallucinate spurious objects, likely from over-compensation for prompt alignment and scene integration. Figures ~\ref{fig:replaceanything_objectignored} and \ref{fig:iclight_objectignored} show cases where the reference object is ignored—a common failure in outpainting/harmonization models that we did not observe with Nova Canvas in our experiments. Figures ~\ref{fig:iclight_productmismatch}, \ref{fig:nova_productmismatch}, and \ref{fig:replaceanything_productmismatch} depict failures caused by misalignment between the ABO category label and the product image; because our captions use the label as the product reference, this mismatch creates ambiguity. The models outcome in this case is unpredictable. Outpainting models (Nova Canvas, Replace Anything) often retain the reference image while also inserting a category-consistent instance or, as in the cup example, produce a semantic compromise. ICLight-being fine-tuned for harmonization—frequently discards the reference entirely. Finally, Figure~\ref{fig:iclight_colorchange} shows an ICLight-specific failure in which objects are recolored to match the scene’s palette.

\FloatBarrier

\FloatBarrier

